\documentclass{article}
\usepackage[table]{xcolor}
\usepackage[preprint]{corl_2026}

\definecolor{sotablue}{HTML}{316EF3}
\newcommand{\sota}[1]{%
  \begingroup\setlength{\fboxsep}{0.8pt}%
  \colorbox{sotablue}{\textcolor{white}{#1}}%
  \endgroup}
\newcommand{\runnerup}[1]{\textbf{#1}}
\newsavebox{\fittablebox}
\newcommand{\fitwidth}[1]{%
  \sbox{\fittablebox}{#1}%
  \ifdim\wd\fittablebox>\linewidth
    \resizebox{\linewidth}{!}{\usebox{\fittablebox}}%
  \else
    \usebox{\fittablebox}%
  \fi}
\usepackage{graphicx}
\usepackage{float}
\usepackage{etoolbox}
\makeatletter
\newlength{\xsparkfloatcskip}
\let\xspark@caption\caption
\newcommand{\xsparkfigcaption}{\par\nointerlineskip\xspark@caption}
\newcommand{\xspark@figsetup}{%
  \setlength{\parskip}{\z@}%
  \setlength{\abovecaptionskip}{\xsparkfloatcskip}%
  \setlength{\belowcaptionskip}{\z@}%
  \let\caption\xsparkfigcaption}
\AtBeginEnvironment{figure}{\xspark@figsetup}
\AtBeginEnvironment{figure*}{\xspark@figsetup}
\renewenvironment{table*}
  {\setlength{\parskip}{\z@}%
   \setlength{\abovecaptionskip}{\z@}%
   \setlength{\belowcaptionskip}{\xsparkfloatcskip}%
   \@dblfloat{table}}
  {\end@dblfloat}

\makeatother
\usepackage{wrapfig}
\usepackage{amsmath}
\usepackage{amssymb}
\usepackage{mathtools}
\usepackage{booktabs}
\usepackage{array}
\usepackage{makecell}

\usepackage{amsmath,amsfonts,bm}

\def\eqref#1{equation~\ref{#1}}
\def\1{\bm{1}}

\DeclareMathAlphabet{\mathsfit}{\encodingdefault}{\sfdefault}{m}{sl}
\SetMathAlphabet{\mathsfit}{bold}{\encodingdefault}{\sfdefault}{bx}{n}

\newcommand{\ie}{\textit{i.e.}, }

\title{Spark-0: Scaling Spatially Aligned Tactile Pretraining for Robot Dexterous Manipulation}

\author{
\begin{minipage}{\dimexpr\textwidth-2\tabcolsep-4pt\relax}
\centering
\normalfont\footnotesize
Tianxing Chen$^{1,3,*,\S}$, Wenwei Lin$^{1,*}$,
Chuqiao Lyu$^{1,*,\S}$, Peicheng Xiang$^{1,*}$, Junyuan Tang$^{4,*}$,\\
Tian Nian$^{1,*}$, Zijian Cai$^{1}$, Kehe Ye$^{1}$,
Chenze Yu$^{1}$, Wenxuan Zhu$^{1}$, Eric J. Chen$^{1}$, Yan Qin$^{4}$, Yue Chen$^{5}$,\\
Tongyu Wu$^{1}$, Junhao Gong$^{1}$, Kai-Chong Lei$^{1}$, Jiahao Zhang$^{1}$, Kailun Su$^{1}$,\\
Qiwei Liang$^{4}$, Renjing Xu$^{4}$,
Ping Luo$^{3,\dag}$, Wenbo Ding$^{1,2,\dag}$\\[0.35em]
\vspace{3pt}
\normalfont\small
$^{1}$Xspark AI \quad
$^{2}$THU \quad
$^{3}$MMLab@HKU \quad
$^{4}$HKUST (GZ) \quad
$^{5}$PKU\\
\normalfont\footnotesize
$^{*}$Equal contribution \quad
$^{\S}$Project leads \quad
$^{\dag}$Corresponding authors\\
\vspace{3pt}
\normalfont\small
\textbf{Project page}: \url{https://xsparkai.com/Spark-0}, \textbf{Code}: \url{https://github.com/XsparkAI/Spark-0}
\end{minipage}
}

\begin{document}

\maketitle

\begin{figure}[H]
  \centering
  \vspace{-2em}
  \includegraphics[width=\textwidth,trim={0 5.5 0 0},clip]{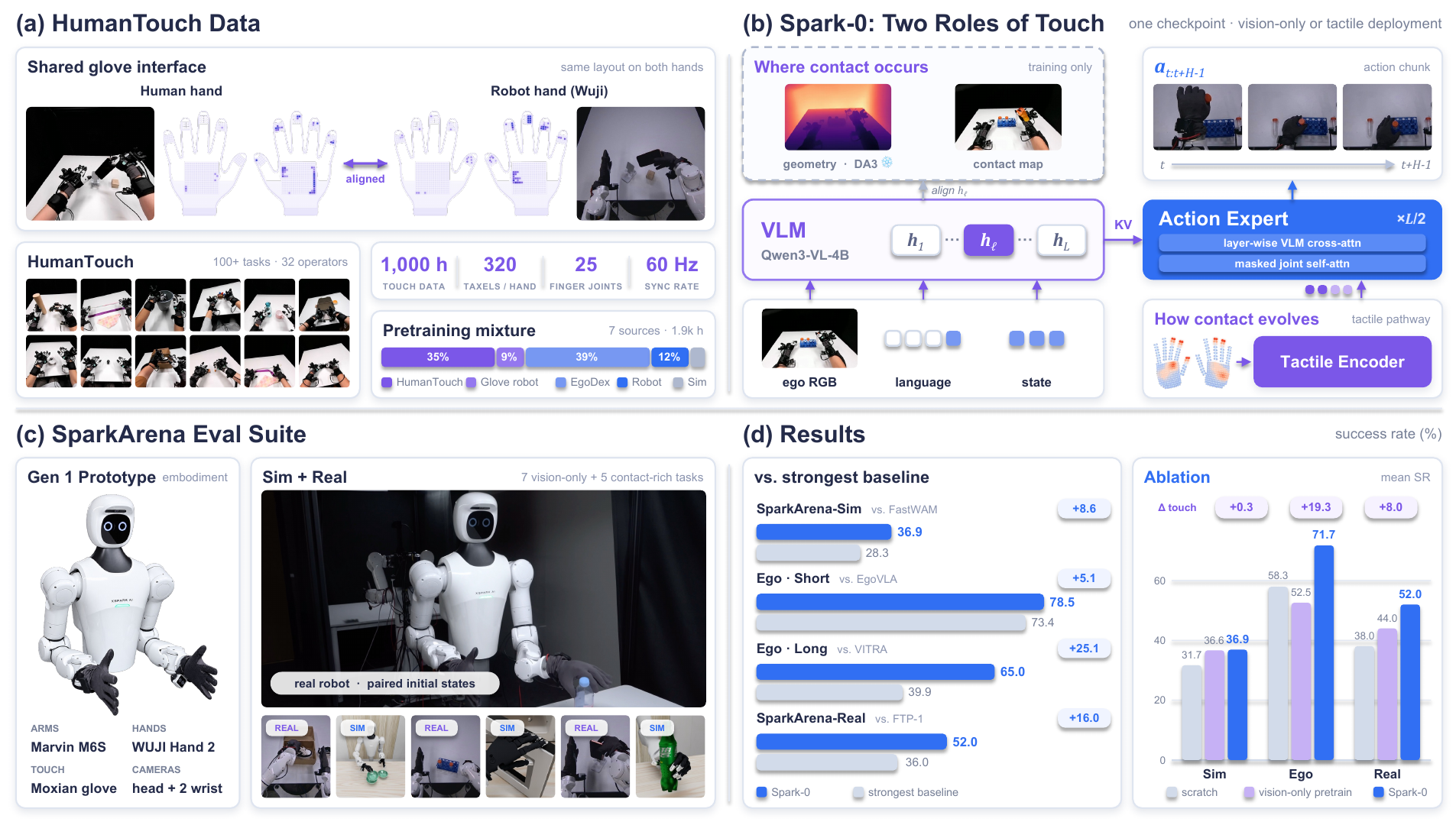}
  % 中文图注：Spark-0 概览。(a) 以 1,000 小时 HumanTouch 加第一视角、机器人和仿真数据预训练。
  % (b) 训练期对齐监督接触发生的位置；触觉编码器以接触动态条件化动作专家。
  % (c) SparkArena：仿真纯视觉任务与 Gen 1 Prototype 真机触觉条件化任务。
  % (d) Spark-0 在 SparkArena 两个赛道及 Ego Humanoid 全部汇总指标上领先。
  \caption{\textbf{Spark-0 overview.}
  \textbf{(a)} Pretraining on $1{,}000$ hours of HumanTouch plus egocentric,
  robot, and simulation data.
  \textbf{(b)} Train-only alignment supervises where contact occurs; a
  tactile encoder conditions the action expert on contact dynamics.
  \textbf{(c)} SparkArena: vision-only tasks in simulation and
  tactile-conditioned tasks on the real Gen 1 Prototype robot.
  \textbf{(d)} Spark-0 leads both SparkArena tracks and all Ego Humanoid
  aggregate metrics.}
  \label{fig:teaser}
\end{figure}

\begin{abstract}
% 中文：通用机器人策略受益于大规模视觉预训练，但在接触密集型灵巧操作中常常表现不佳，
% 因为力、柔顺性和初始滑移难以从图像中观测。我们提出 Spark-0，一个空间对齐的视觉-触觉-
% 语言-动作（VTLA）模型，以两种互补方式使用触觉。预训练中，由触觉手套生成的图像空间
% 接触图与冻结 3D 几何教师的特征共同监督视觉 token，这些目标在推理时移除；同时，触觉
% 通路编码区域力历史与 3D taxel 状态，条件化动作专家。因此单一 checkpoint 同时支持纯视觉
% 与触觉条件化部署。为规模化这种监督，我们构建 1,000 小时全手触觉人类操作数据集
% HumanTouch，并与第一视角视频、机器人数据和仿真共同预训练。我们还开源 SparkArena，在同一
% 灵巧手本体上配对纯视觉仿真任务与触觉条件化真机任务。Spark-0 在两条 SparkArena 轨道上
% 取得最高平均成功率，并在 Ego Humanoid 全部八项汇总指标上排名第一。与使用相同数据与计算量、
% 仅舍弃触觉的纯视觉预训练相比，保留 1,000 小时 HumanTouch 触觉使 Ego Humanoid 纯视觉成功率从
% 52.5% 提升至 71.7%，使部署时同样使用触觉的 SparkArena-Real 成功率从 44.0% 提升至 52.0%，表明触觉不仅
% 是策略输入，也是灵巧视觉运动预训练的可规模化监督信号。
Generalist robot policies benefit from large-scale visual pretraining but
often struggle with contact-rich dexterous manipulation, where force,
compliance, and incipient slip are hard to observe from images. We present
\emph{Spark-0}, a spatially aligned vision-tactile-language-action (VTLA)
model that uses touch in two complementary ways. During pretraining,
image-space contact maps derived from tactile gloves, together with features
from a frozen 3D geometry teacher, supervise the visual tokens; these
objectives are removed at inference. In parallel, a tactile pathway encodes
regional force histories and 3D taxel states to condition the action expert.
A single checkpoint therefore supports both vision-only and
tactile-conditioned deployment. To scale this supervision, we build
\emph{HumanTouch}, a $1{,}000$-hour dataset of full-hand tactile human
manipulation, and pretrain Spark-0 on it together with egocentric video,
robot data, and simulation. We also release \emph{SparkArena}, which pairs
vision-only simulated tasks with tactile-conditioned real-robot tasks on one
dexterous embodiment. Spark-0 achieves the highest mean success rate on both
SparkArena tracks and ranks first on all eight aggregate metrics of the Ego
Humanoid Manipulation Benchmark. Compared with vision-only pretraining on
the same data and compute with touch discarded, retaining the touch of
$1{,}000$ hours of HumanTouch raises vision-only success on Ego Humanoid
from $52.5\%$ to $71.7\%$, and success on SparkArena-Real, where
touch is also used at deployment, from $44.0\%$ to $52.0\%$, indicating
that touch serves not only as a policy input but also as scalable
supervision for dexterous visuomotor pretraining.
\end{abstract}

\keywords{tactile pretraining, dexterous manipulation, robot learning}

\section{Introduction}

% 中文：大规模视觉预训练赋予 VLA 模型强大的语义与视觉运动先验，但接触密集型灵巧操作
% 仍然困难：力、柔顺性和初始滑移难以从图像中观测，且手部常遮挡接触区域。已有工作通过
% 时序触觉反馈、通用触觉预训练、人到机器人触觉迁移以及预测潜在触觉 token 的视触预训练
% 引入触觉。仍存在三个缺口：缺少与手部姿态和多视角视频同步的大规模全手触觉数据；触觉
% 通常只用于条件化或预测动作，而非监督视觉表征；很少有基准同时衡量视觉灵巧性与接触
% 密集型触觉控制。因此我们提出：触觉能否作为视觉运动预训练的可规模化监督，而不仅是
% 策略的附加输入？
Large-scale visual pretraining gives vision-language-action (VLA) models
strong semantic and visuomotor priors
\cite{intelligence2025pi_,bjorck2025gr00t,community2026starvla}, yet
contact-rich dexterous manipulation remains difficult: force, compliance, and
incipient slip are weakly observable from images, and the hand often
occludes the contact region. Prior work adds touch through temporal tactile
feedback \cite{trex}, generalist tactile pretraining \cite{yuan2026ftp},
human-to-robot tactile transfer \cite{ttp}, and visuo-tactile pretraining
with predicted latent tactile tokens \cite{n0vtla}. Three gaps remain.
Large-scale full-hand tactile data synchronized with hand pose and
multi-view video are scarce; touch is typically used to condition or predict
actions rather than to supervise visual representations; and few benchmarks
measure both visual dexterity and contact-rich tactile control. We therefore
ask: \emph{can touch serve as scalable supervision for visuomotor
pretraining, not only as an additional policy input?}

% 中文：回答这一问题首先需要规模化的触觉数据。机器人遥操作成本高且接触覆盖有限，因此我们
% 转向人类示范，构建 HumanTouch：1,000 小时操作数据，覆盖 100 余项任务和 32 名操作者，以
% 60 Hz 同步记录双手全手触觉手套、手部与手腕追踪以及头部和腕部相机。同款手套可安装于目标
% 灵巧手，使人类与机器人数据共享触觉布局。
Answering this question first requires tactile data at scale. Robot
teleoperation is costly and covers a narrow range of contacts, so we turn to
human demonstrations and build \textbf{HumanTouch}: $1{,}000$ hours of
manipulation across more than $100$ tasks and $32$ operators, recorded with bilateral full-hand tactile gloves, hand and wrist
tracking, and head and wrist cameras synchronized at $60$~Hz. The same
gloves can be mounted on the target dexterous hand, giving human and robot
data a shared tactile layout.

% 中文：基于这些数据，我们训练空间对齐的 VTLA 模型 Spark-0。它不仅将触觉作为策略输入，
% 还用触觉监督视觉表征，赋予触觉两种作用（图~\ref{fig:teaser}）。其一，由手套生成并渲染到头部
% 视角的接触图与冻结 3D 几何教师的特征共同监督视觉 token，提供接触发生位置的线索；这些
% 目标仅用于训练，因此部署时即使没有触觉传感器，模型也能受益于触觉。其二，触觉通路
% 编码区域力历史与 3D taxel 状态，只条件化动作专家，告诉它接触如何演化。原始触觉不进入
% 视觉-语言骨干，单一 checkpoint 同时支持纯视觉与触觉条件化部署。
With this data, we train \textbf{Spark-0}, a spatially aligned
vision-tactile-language-action (VTLA) model that uses touch to supervise
visual representations as well as to condition actions, assigning it two
roles (Figure~\ref{fig:teaser}). First,
glove-derived contact maps rendered into the head view, together with
features from a frozen 3D geometry teacher, supervise the visual tokens with
cues about \emph{where contact occurs}. These objectives are train-only, so
the model can benefit from touch even when no tactile sensor is available at
deployment. Second, a tactile pathway encodes regional force histories and 3D
taxel states and conditions only the action expert on \emph{how contact
evolves}. Raw touch never enters the vision-language backbone, so one
checkpoint serves both vision-only and tactile-conditioned deployment.

% 中文：评测这两种作用需要在同一本体上分别测试有触觉和无触觉的部署，现有基准无法满足。
% 因此我们开源 SparkArena，并发布全部任务定义、资产、成对初始条件和评分协议。在同一灵巧手本体上，SparkArena-Sim 测试纯视觉灵巧操作，SparkArena-Real 测试
% 使用在线触觉的接触密集型操作。为检验无测试时触觉的跨本体迁移，我们还在 Ego Humanoid
% 上评测。Spark-0 在两条 SparkArena 轨道上取得最高平均成功率，并在 Ego Humanoid 全部八项
% 汇总指标上排名第一。与使用相同数据与计算量、仅舍弃触觉的纯视觉预训练相比，保留
% 1,000 小时 HumanTouch 触觉使 Ego Humanoid 纯视觉成功率从 52.5% 提升至 71.7%，使 SparkArena-Real 从 44.0% 提升至 52.0%。
Evaluating these two roles requires testing deployment both with and without
touch on the same embodiment, which existing benchmarks do not support. We
therefore release \textbf{SparkArena}, with all task definitions, assets,
paired initial conditions, and scoring protocols.
On the same dexterous embodiment, SparkArena-Sim tests vision-only dexterity
and SparkArena-Real tests contact-rich manipulation with online touch. To
test cross-embodiment transfer without test-time touch, we additionally
evaluate on the Ego Humanoid Manipulation Benchmark \cite{yang2025egovla}.
Spark-0 achieves the highest mean success rate on both SparkArena tracks and
ranks first on all eight Ego Humanoid aggregate metrics. Compared with
vision-only pretraining on the same data and compute with touch discarded,
retaining the touch of $1{,}000$ hours of HumanTouch raises vision-only
success on Ego Humanoid from $52.5\%$ to $71.7\%$; on
SparkArena-Real, the same comparison raises success from $44.0\%$ to
$52.0\%$.

\paragraph{Contributions}
\begingroup
\setlength{\leftmargini}{1.25em}
\begin{itemize}
    % 中文：HumanTouch：1,000 小时全手触觉人类操作数据集，采集手套也可安装于机器人手。
    \item \textbf{HumanTouch}, a $1{,}000$-hour full-hand tactile dataset of
    human manipulation, collected with gloves that also mount on the robot
    hand.

    % 中文：Spark-0：将触觉同时用作视觉 token 的仅训练监督和动作专家输入的 VTLA 模型，
    % 单一 checkpoint 支持纯视觉与触觉条件化部署。
    \item \textbf{Spark-0}, a VTLA model that uses touch both as train-only
    supervision for visual tokens and as an input to the action expert,
    supporting vision-only and tactile-conditioned deployment from one
    checkpoint.

    % 中文：SparkArena：在同一灵巧手本体上配对纯视觉仿真任务与触觉条件化真机任务的开源基准。
    \item \textbf{SparkArena}, an open benchmark that pairs vision-only
    simulated tasks with tactile-conditioned real-robot tasks on one
    dexterous embodiment.

    % 中文：实验：Spark-0 在三项基准上优于强基线，且大规模触觉预训练显著提升未见本体上的
    % 纯视觉性能。
    \item \textbf{Experiments} on three benchmarks showing that Spark-0
    outperforms strong baselines and that large-scale tactile pretraining
    substantially improves vision-only performance on an unseen embodiment.
\end{itemize}
\endgroup

\section{Related Work}

\subsection{Tactile data}

% 中文：触觉传感器包括高分辨率局部传感器和全手柔性阵列。现有数据来自人类交互、视触
% 夹爪和仿真，表征学习工作探索触觉估计、生成、补全、跨传感器聚合以及人类触觉迁移。
% 这些数据大多规模有限或缺少同步的手部姿态与多视角视频；HumanTouch 以 1,000 小时规模
% 提供二者，且采集手套由人类与机器人共享。
Tactile sensors range from high-resolution local sensors to full-hand
flexible arrays \cite{gelsight,digit}. Existing data come from human
interaction, visuo-tactile grippers, and simulation
\cite{opentouch,worldinyourhands,freetacman,touchinthewild,chen2026univtac},
and representation learning explores tactile estimation, generation,
completion, cross-sensor aggregation, and human-to-robot transfer
\cite{zhou2026touchanything,zhou2026touchsight,liang2026biview,yuan2026ftp,n0foundation,ttp}.
Most of these datasets are limited in scale or lack synchronized hand pose
and multi-view video. HumanTouch provides both at $1{,}000$ hours, with
gloves shared by human and robot acquisition.

\subsection{Policies for contact-rich dexterous manipulation}

% 中文：VLA 模型以扩散或 flow-matching 动作专家生成动作块，统一动作空间、3D 语义物体表征、第一视角视频、
% 人手动作迁移和世界动作模型扩大了视觉与运动覆盖；其中记录触觉的方法也只将其作为策略输入，而非视觉表征的预训练信号。触觉策略将触觉用于
% 动作条件化、预测和反馈。最相关的 STAR 以 200 小时双臂真机数据处理触觉稀疏性；
% Spark-0 则结合 HumanTouch，在预训练中用触觉同时监督视觉表征并条件化动作专家。
VLA models extend vision-language models to robot control
\cite{zitkovich2023rt,kim2024openvla} with diffusion or flow-matching action
experts
\cite{zhao2023learning,chi2025diffusion,black2024pi_0,intelligence2025pi_,bjorck2025gr00t,community2026starvla,kim2026rldx}.
Unified action spaces \cite{liu2024rdt,zheng2025x}, 3D semantic object
representations \cite{Chen_2025_CVPR}, egocentric video
\cite{egodex,hyvla}, human hand-motion transfer
\cite{dexumi,yang2025egovla,li2026scalable,bi2025h,chen2026unit}, and world
action models \cite{yuan2026fast} broaden visual and motor coverage; where
these methods record touch, it serves as a policy input rather than as a
pretraining signal for visual representations.
Tactile policies use touch for action conditioning,
prediction, or feedback
\cite{trex,n0vtla,n0twam,qin2026dextouch,lift,tacforcing,yuan2026ftp}.
The closest work, STAR, handles tactile sparsity with 200 hours of bimanual
robot data \cite{liu2026star}. Spark-0 instead pretrains on HumanTouch and
heterogeneous data, using touch both to supervise visual representations and
to condition the action expert.

\subsection{Manipulation benchmarks}

% 中文：经典基准与双臂、记忆、仿真-真机基准主要面向夹爪；灵巧与视触基准大多仅在仿真
% 中评测。SparkArena 在同一灵巧手本体上同时评测纯视觉灵巧操作与真机触觉控制。
Classic and recent benchmarks mainly evaluate gripper-based multi-task,
long-horizon, bimanual, memory-dependent, or sim-and-real manipulation
\cite{james2020rlbench,mees2022calvin,liu2023libero,chen2025robotwin,chen2026rmbench,chen2026robodojo}.
Dexterous and visuo-tactile benchmarks
\cite{wang2026dexjoco,yao2026dexverse,yang2025egovla,chen2026univtac} mostly
evaluate in simulation. SparkArena instead evaluates both vision-only
dexterity and real-robot tactile control on the same dexterous embodiment.

\section{HumanTouch and Pretraining Data}
\label{sec:humantouch}

\begin{figure*}[h]
  \centering
  \includegraphics[width=\textwidth]{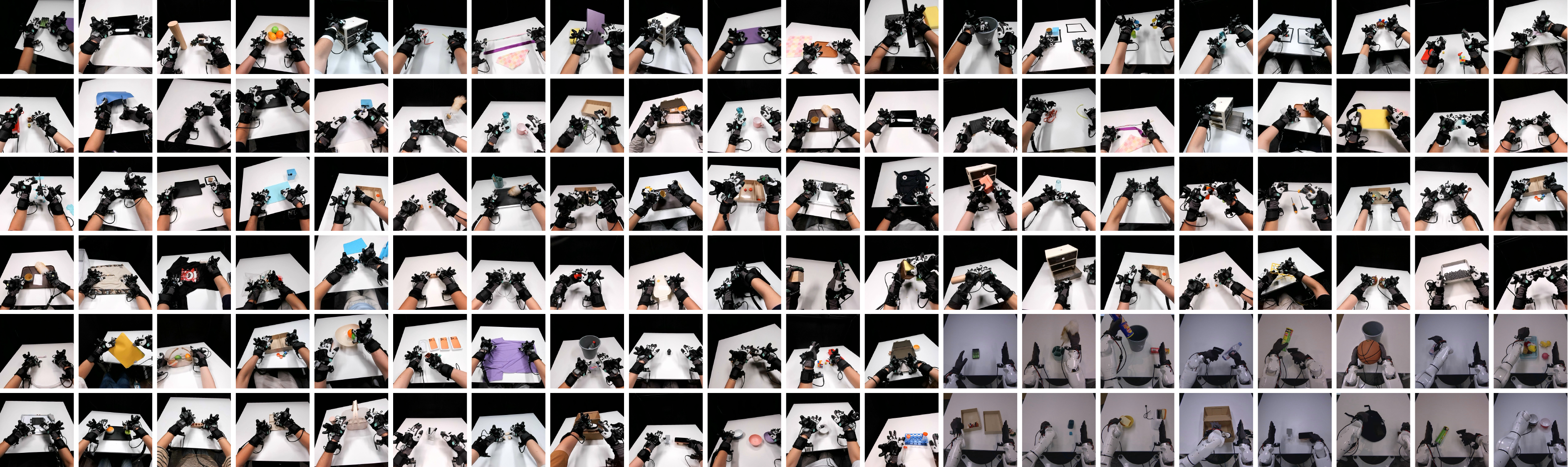}
  % 中文图注：HumanTouch 概览：可穿戴视触采集下的代表性任务与场景。
  \caption{\textbf{HumanTouch overview.} Representative in-the-wild tasks
  spanning diverse objects, scenes, and motions.}
  \label{fig:dataset-overview}
\end{figure*}

\subsection{Acquisition hardware}

% 中文：每套采集设备包含双手柔性压阻式手套（每只手 460 个感知点）、25 关节手指追踪、
% 6-DoF 手腕追踪、头部场景相机和两个腕部相机，均以 60 Hz 同步。我们屏蔽每只手位于弯折
% 区域、易因弯曲产生伪压力的 140 个感知点，其余 320 个既作为触觉输入，也用于生成接触图。
% 操作者无需机器人即可完成自然的接触密集型任务；同款手套也可安装于目标灵巧手，以相同的
% 信号布局记录机器人示范。
Each acquisition unit combines bilateral flexible piezoresistive gloves with
$460$ sensing points per hand, 25-joint finger tracking, 6-DoF wrist
tracking, a head-mounted scene camera, and two wrist-mounted cameras, all
synchronized at $60$~Hz (Appendix Figure~\ref{fig:humantouch-hardware}). We
mask the $140$ points per hand that lie over flexion regions, where bending
produces spurious pressure, and use the remaining $320$ both as tactile
input and to generate contact maps (Appendix
Figure~\ref{fig:glove-taxel-mask}). Operators perform natural contact-rich
tasks without a robot in the loop, and the same gloves can be mounted on
the target dexterous hand to record robot demonstrations with the same
signal layout.

\subsection{Dataset scale and diversity}

% 中文：HumanTouch 包含 1,000 小时多模态操作数据，覆盖 100 余项任务和 32 名操作者。
% 任务在接触模式、物体属性、动作复杂度和 episode 时长上具有多样性，而非重复少量动作
% （图~\ref{fig:dataset-overview}）。数据细节见附录。
HumanTouch contains $1{,}000$ hours of multimodal manipulation data spanning
more than $100$ tasks and $32$ operators. Tasks vary in contact pattern,
object properties, motion complexity, and episode duration rather than
repeating a small set of motions (Figure~\ref{fig:dataset-overview}).
Further details are given in Appendix~\ref{app:humantouch_details}.

\subsection{Pretraining mixture}
\label{sec:data_mixture}

% 中文：Spark-0 在异构数据混合上预训练，各来源作用不同。HumanTouch 提供人类示范中的
% 全手接触监督；EgoDex 第一视角视频扩展场景、物体与动作覆盖；真机数据对齐目标本体并
% 增加机器人任务多样性，其中部分数据使用同款手套记录；域随机化仿真增加视觉变化。各来源
% 时长、采样权重、相机配置与 400k 步训练设置见附录。
Spark-0 is pretrained on a heterogeneous mixture in which each source serves
a distinct purpose. HumanTouch provides full-hand contact supervision from
human demonstrations; EgoDex egocentric video \cite{egodex} broadens scene,
object, and motion coverage; real-robot data align the model with the target
embodiment and add robot-task diversity, with a subset recorded through the
same gloves; and domain-randomized simulation adds visual variation. Source
hours, sampling weights, camera configurations, and the 400k-step schedule
are given in Appendix~\ref{app:pretraining_recipe}.

\section{Spark-0}
\label{sec:method}

\begin{figure*}[t]
  \centering
  \includegraphics[width=\textwidth,trim={0 5 0 0},clip]{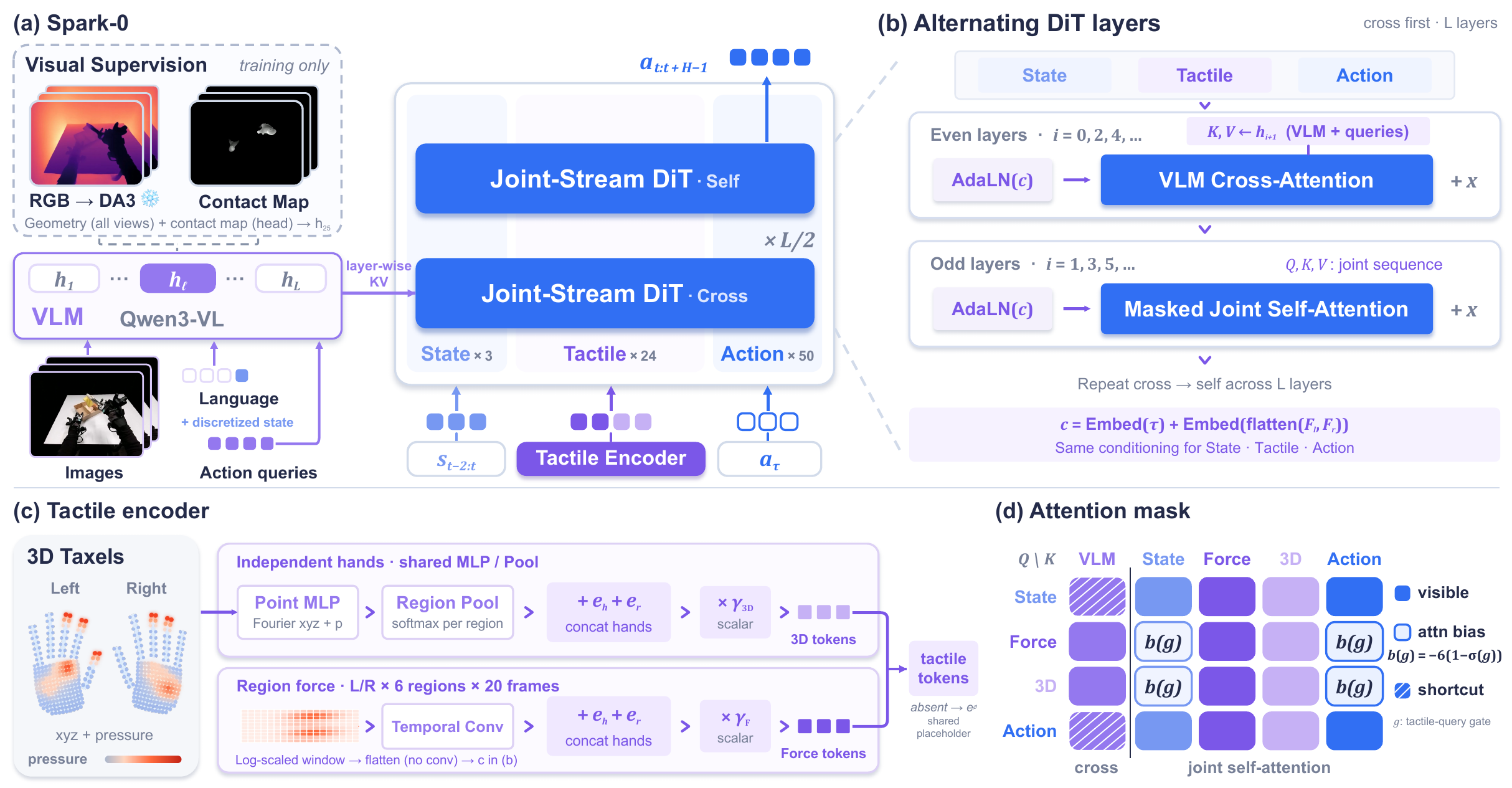}
  % 中文图注：Spark-0 架构。(a) Qwen3-VL 为 flow-matching 动作专家提供上下文；仅训练
  % 阶段的目标对齐视觉 token，原始触觉不进入 VLM。(b) 偶数层交叉注意 VLM，奇数层做
  % State--Tactile--Action 联合自注意力，均使用力条件化 AdaLN。(c) 区域力与 3D taxel
  % （每只手 320 个）两路共 24 个触觉 token。(d) VLM 列为偶数层交叉注意力（含训练捷径），其余列为奇数层
  % 自注意力。细节见附录。
  \caption{\textbf{Spark-0 architecture.}
  \textbf{(a)} Qwen3-VL conditions a flow-matching action expert; train-only
  objectives align visual tokens, and raw touch bypasses the VLM.
  \textbf{(b)} Even layers cross-attend to the VLM; odd layers self-attend
  over State--Tactile--Action tokens. All layers use force-conditioned AdaLN.
  \textbf{(c)} Regional force and 3D taxel ($320$ per hand) streams yield
  $24$ tactile tokens.
  \textbf{(d)} The VLM column is even-layer cross-attention, including the
  training shortcut; the other columns are odd-layer self-attention.
  Details are in Appendix~\ref{app:architecture_details}.}
  \label{fig:spark0_architecture}
\end{figure*}

% 中文：接触密集型灵巧控制需要两类视觉难以提供的信息：接触发生在哪里，以及接触如何演化。
% 前者部分可见，但语义化的 VLM 特征并不显式编码；后者涉及被遮挡的压力变化。Spark-0 分别
% 处理二者：仅训练的空间对齐用接触区域监督视觉 token（第~\ref{sec:spatial_alignment}节），
% 触觉通路以接触动态条件化动作专家（第~\ref{sec:tactile_fusion}节）。我们先介绍骨干与动作空间。
Contact-rich dexterous control needs two kinds of information that vision
provides poorly: \emph{where} contact occurs and \emph{how} it evolves. The
former is partly visible but not explicitly encoded by semantic VLM
features; the latter involves occluded pressure changes. Spark-0 addresses
them separately: train-only spatial alignment supervises visual tokens with
contact regions (Section~\ref{sec:spatial_alignment}), and a tactile pathway
conditions the action expert on contact dynamics
(Section~\ref{sec:tactile_fusion}). We first describe the backbone and
action space.

\subsection{VLA backbone and action objective}
\label{sec:vla_backbone}

% 中文：Spark-0 基于 StarVLA 与 XPolicyLab 代码库，将 Qwen3-VL-4B 与逐层 flow-matching
% 动作专家结合，使动作生成可利用从低层空间特征到高层语义的完整视觉层级。给定图像、指令和
% 机器人状态，VLM 输出各层隐状态。当前状态以离散 token 进入 VLM 序列，状态历史以连续形式
% 嵌入动作专家。追加在 VLM 序列末尾的可学习 action query 汇总视觉-语言上下文，因果掩码使其
% 不改变前序 token。
Spark-0 builds on StarVLA \cite{community2026starvla} and the XPolicyLab
codebase \cite{community2026xpolicylab}. It pairs Qwen3-VL-4B with a
layerwise flow-matching action expert so that action generation can draw on
the full visual hierarchy, from low-level spatial features to high-level
semantics. Given images $I$, instruction $w$, and robot state $s$, the VLM
produces hidden states $\{h^{(l)}\}_{l=1}^{L}$. The current state enters the
VLM sequence as discretized tokens, and the state history is embedded
continuously in the action expert. Learnable action queries appended to the VLM sequence
summarize the visual-language context; causal masking prevents them from
altering earlier tokens.

% 中文：统一状态与动作空间。各数据源的手部形态和坐标系不同，因此全部离线转换为目标本体的
% 双手布局：每侧 $k$ 在时刻 $t$ 有末端位姿 $T^k_t$ 和 WUJI Hand 2 的 20 个关节角 $q^k_t$。
% 人类数据将手腕轨迹配准到机器人坐标系，并通过逆运动学将手部关键点重定向到 WUJI 手。
% $\rho$ 将位姿编码为平移与由旋转矩阵前两行组成的 6D 表示，$\rho^{-1}$ 通过正交化恢复
% 旋转。状态使用头部相机坐标系下的绝对位姿，动作块第 $h$ 行使用相对当前末端的位姿与
% 绝对关节目标。相机坐标系为所有数据源提供以视觉为中心的共同参考系，相对动作块在每次
% 推理时重新锚定到最新观测位姿。状态与动作均为 58 维。
\paragraph{Unified state and action space}
Human and robot sources differ in hand morphology and coordinate frames, so
all sources are converted offline to the bimanual layout of the target
embodiment: for side $k\in\{\mathrm{L},\mathrm{R}\}$ at time $t$, an
end-effector pose $T^k_t\in\mathrm{SE}(3)$ and the $20$ joint angles $q^k_t$
of the WUJI Hand 2. For human data (HumanTouch and EgoDex), wrist
trajectories are registered to the robot frame and hand keypoints are
retargeted to the WUJI hand by inverse kinematics. Let $C_t$ be the
head-camera pose, and let $\rho(T)=[\mathbf{p},\,r(R)]\in\mathbb{R}^9$
encode a pose by its translation $\mathbf{p}$ and a 6D rotation $r(R)$
formed from the first two rows of $R$; $\rho^{-1}$ recovers a valid
rotation by orthonormalization. States use absolute poses in the
head-camera frame, and the $h$-th action row uses the pose relative to the
current end effector, with absolute joint targets:
\begin{equation}
  s^k_t=\bigl[\rho\bigl(C_t^{-1}T^k_t\bigr),\;q^k_t\bigr],
  \qquad
  a^k_{t+h}=\bigl[\rho\bigl((T^k_t)^{-1}T^k_{t+h}\bigr),\;q^k_{t+h}\bigr],
  \quad h=0,\dots,H-1.
  \label{eq:action_space}
\end{equation}
The camera frame gives all sources a shared, vision-centric reference, and
the relative chunk is re-anchored to the latest observed pose at each
inference step; deployment recovers absolute targets as
$T^k_{t+h}=T^k_t\,\rho^{-1}(\hat a^k_{t+h})$ from the predicted pose
component $\hat a^k_{t+h}$. Concatenating both sides yields
$2\times(9+20)=58$-dimensional states and actions.

% 中文：动作专家包含 36 个宽度 1024 的 DiT block，与 VLM 层数一致。偶数 block 交叉注意
% 对应 VLM 层的投影隐状态（含 64 个 action query），奇数 block 执行联合自注意力。专家根据
% 三帧状态预测 50 步动作块，以 flow matching 训练；损失对每个样本的 6 次噪声采样取平均，
% 推理时以 4 步 Euler 积分生成动作。
\paragraph{Action expert}
The action expert has $L=36$ DiT blocks of width $1024$, matching the VLM
depth. Even blocks cross-attend to the projected hidden state of the
corresponding VLM layer, including $64$ action queries; odd blocks perform
the joint self-attention of Section~\ref{sec:tactile_fusion}. From three
state frames, the expert predicts an $H=50$ action chunk $a=a_{t:t+H-1}$.
With normalized target chunk $a$, Gaussian noise $\epsilon$, and flow time
$\tau\in[0,1]$, it is trained with
\begin{equation}
  a^{\tau}=(1-\tau)\epsilon+\tau a,
  \qquad
  \mathcal{L}_{\mathrm{act}}
  =
  \mathbb{E}_{a,\tau,\epsilon}
  \left[
    \left\|
      v_\theta(a^{\tau},\tau,\mathcal{C})-(a-\epsilon)
    \right\|_2^2
  \right],
  \label{eq:flow_loss}
\end{equation}
where $\mathcal{C}$ denotes the continuous state, layerwise VLM context, and
optional tactile tokens. The loss is averaged over six noise draws per
example, and inference integrates the predicted velocity with four Euler
steps. Normalization and flow-time sampling are given in
Appendix~\ref{app:pretraining_recipe}.

\subsection{Spatial alignment: supervising interaction regions}
\label{sec:spatial_alignment}

% 中文：动作专家逐层读取 VLM 特征，因此我们希望这些特征编码交互发生的位置。我们在 VLM
% 第 25 层施加两个仅训练目标；该层较深但非最深，且被一个交叉注意力 block 读取。几何对齐
% 为所有视角和数据源提供稠密场景结构，接触对齐在有手套数据时标出交互区域。两个预测头在
% 推理时移除。
Because the action expert reads VLM features layer by layer, we want those
features to encode where interaction occurs. We apply
two train-only objectives to VLM layer $\ell^\star=25$, a deep but not final
layer that is read by a cross-attention block. Geometry alignment provides
dense scene structure for every view and source; contact alignment marks
interaction regions whenever glove data are available. Both heads are
discarded at inference.

% 中文：几何对齐。冻结的 Depth-Anything-3 教师产生的 patch 特征被重采样到合并后的视觉
% token 网格，与每个视觉 token 一一对应，并以余弦损失与投影后的 VLM token 对齐。损失先在
% 视角内平均、再在视角间平均，避免 token 更多的相机主导目标。
\paragraph{Geometry alignment}
A frozen Depth-Anything-3 teacher \cite{da3} produces patch features
resampled onto the merged visual-token grid, so that
$z^{\mathrm{geo}}_{b,v,j}$ matches visual token $j$ in view $v$. We align
them with projected VLM tokens via
\begin{equation}
  \mathcal{L}_{\mathrm{geo}}
  =
  \mathbb{E}_{b,v,j}
  \left[
    1-\cos\!\left(
      \phi_{\mathrm{geo}}(h^{(\ell^\star)}_{b,v,j}),
      z^{\mathrm{geo}}_{b,v,j}
    \right)
  \right].
  \label{eq:geo_alignment}
\end{equation}
The loss is averaged within each view and then across views, so that
cameras with more tokens do not dominate.

% 中文：接触对齐。几何描述场景，但无法判断手是否真正接触物体。对每只手，我们检测处于接触
% 中的 taxel，将其落到由手套重建的 MANO 网格上，沿手表面扩散强度，并把网格渲染到头部视角；
% 双手取逐像素最大值。热图池化到 token 网格作为软标签，轻量预测头从头部视角 token 预测
% 接触 logit。由于接触区域稀疏，我们组合 focal 与 soft Dice 损失。接触图只作为训练目标，
% 从不作为策略输入，构建细节见附录。
\paragraph{Contact alignment}
Geometry describes the scene but cannot tell whether the hand touches an
object. For each hand, we detect taxels in contact, anchor them to a MANO
mesh reconstructed from the glove, diffuse their intensity along the hand
surface, and render the mesh into the head view; the two hands are merged by
a pixelwise maximum. The map is pooled to the token grid as a soft target
$y_{b,j}\in[0,1]$, and a lightweight head predicts contact logits from the
head-view tokens $h^{(\ell^\star)}_{b,0,j}$. Because contact regions are
sparse, we combine focal and soft Dice losses:
\begin{equation}
  \mathcal{L}_{\mathrm{con}}
  =
  0.1\mathcal{L}_{\mathrm{focal}}
  +0.1\mathcal{L}_{\mathrm{dice}}.
  \label{eq:contact_alignment}
\end{equation}
The contact map is a training target only, never a policy input; its
construction is detailed in Appendix~\ref{app:contact_maps}.

\subsection{Tactile fusion: conditioning actions on contact dynamics}
\label{sec:tactile_fusion}

% 中文：空间对齐改善视觉可推断的信息，但无法恢复被遮挡的压力变化。因此我们加入只接入
% 动作专家的触觉通路；原始触觉不进入 VLM，纯视觉部署时以可学习占位符替代触觉 token。
% 接触包含两种时间尺度：压力在毫秒级变化，接触几何则可由单帧描述。因此使用两个编码器：
% 时序卷积汇总每只手六个区域 20 帧的力历史，得到 12 个区域力 token；共享点编码器嵌入当前
% 帧每只手 320 个 taxel（头部相机坐标系 xyz 与压力），并在相同区域内做注意力池化，得到 12 个
% 3D token。
Spatial alignment improves what vision can infer but cannot recover occluded
pressure changes. We therefore add a tactile pathway that feeds only the
action expert; raw touch never enters the VLM, and vision-only deployment
replaces tactile tokens with learned placeholders. Contact evolves on two
time scales: pressure changes within milliseconds, whereas contact geometry
is captured by a single frame. A temporal convolution therefore summarizes a
$20$-frame ($\approx333$~ms) force history for each of six regions per hand,
yielding $12$ regional force tokens, and a shared point encoder embeds the
current $320$ taxels per hand (head-camera-frame xyz and pressure) and
attention-pools them within the same regions, yielding $12$ 3D tokens.

% 中文：缺失触觉。预训练数据中只有部分样本带有触觉，而传感器缺失不同于零压力。存在的
% 触觉流由各自独立的可学习门控缩放，缺失或无效的流使用共享可学习占位符，使纯视觉与触觉
% 样本可以混合成批训练，而不会把缺失当作无接触。
\paragraph{Missing touch}
Only part of the pretraining data carries touch, and a missing sensor
differs from zero pressure. Present streams are scaled by independent
learned gates, while absent or invalid streams use a shared learned
placeholder $e^{\varnothing}$. This fixed layout lets vision-only and
tactile samples share a batch without treating missing sensing as absent
contact. Encoder details, pressure conditioning, quality checks, and gate
initialization are given in Appendix~\ref{app:architecture_details}.

% 中文：动作-触觉联合序列。接触应随动作计划共同演化，而非作为固定外部上下文，因此触觉
% token 与 State、Action token 组成同一序列，共 3 个 State、12 个区域力、12 个 3D 和 50 个
% Action token。交叉注意力层让所有行读取视觉-语言上下文，自注意力层联合更新各行。非对称
% 偏置仅限制触觉 query 读取 State 与 Action，使触觉 token 保持以传感为主。动作专家可能只
% 依赖视觉-语言通路，因此训练时以低概率将触觉样本中 State 与 Action 行的交叉注意力输出
% 置零，迫使上下文经由触觉行传递。区域力窗口还通过零初始化 AdaLN 调制每一层。
\paragraph{Joint action-tactile stream}
Contact should evolve with the action plan rather than serve as fixed
external context, so tactile tokens join the State and Action tokens in one
sequence:
\begin{equation}
  Z^{(0)}_b
  =
  \left[
    E_s(s_b)
    \ \Vert\
    \tilde{e}^{\mathrm{reg}}_b
    \ \Vert\
    \tilde{e}^{\mathrm{3d}}_b
    \ \Vert\
    E_a(a^{\tau}_b)
  \right],
  \label{eq:joint_sequence}
\end{equation}
where $E_s$ and $E_a$ embed each 58-dimensional state frame and noisy action
row into width $1024$, giving three State, twelve regional force, twelve 3D,
and fifty Action tokens. Cross-attention blocks let all rows read the
visual-language context, and self-attention blocks update them jointly. To
keep tactile tokens grounded in sensing, an asymmetric bias limits only how
tactile queries attend to State and Action keys; State and Action attend to
the gated tactile tokens without bias. Because the action expert could
otherwise rely solely on the visual-language path, training on tactile
samples occasionally zeros the cross-attention outputs of the State and
Action rows, forcing context to flow through the tactile rows. The regional
force window also modulates every block through a zero-initialized AdaLN
pathway, and actions are decoded from the final $H$ Action rows. Details are
in Appendix~\ref{app:architecture_details}.

\subsection{Training objective and deployment}
\label{sec:joint_objective}

% 中文：所有组件以如下目标联合训练。几何对齐作用于所有可用视角，接触对齐作用于有效接触图，
% 触觉编码器在含触觉样本上通过动作损失学习。推理时移除 DA3 教师与对齐预测头，并关闭触觉
% 捷径，非对称偏置与门控触觉通路保留。同一组权重既可填入触觉占位符纯视觉部署，也可使用
% 在线触觉信号部署，无需额外模块。
All components are trained jointly with
\begin{equation}
  \mathcal{L}
  =
  \mathcal{L}_{\mathrm{act}}
  +
  \lambda_{\mathrm{geo}}\mathcal{L}_{\mathrm{geo}}
  +
  \mathcal{L}_{\mathrm{con}},
  \qquad
  \lambda_{\mathrm{geo}}=0.2.
  \label{eq:total_loss}
\end{equation}
Geometry alignment applies to every available view, contact alignment to
valid contact maps, and the tactile encoders learn through
$\mathcal{L}_{\mathrm{act}}$ on samples with touch. At inference, the DA3
teacher and alignment heads are removed and the tactile shortcut is
disabled, while the asymmetric bias and gated tactile pathway remain. The
same weights support vision-only deployment with tactile placeholders and
tactile-conditioned deployment with online signals, without additional
modules.

\section{SparkArena Evaluation Suite}
\label{sec:sparkarena}

% 中文：Spark-0 以两种方式使用触觉，因此评测需回答两个问题：触觉预训练能否在无触觉时改善
% 视觉策略，在线触觉能否改善接触密集型控制。SparkArena 为每个问题提供一条轨道。两条轨道
% 使用同一 Gen 1 Prototype 双臂灵巧手本体，保持机器人形态与相机布局一致。真机安装与 HumanTouch 相同
% 布局的 Moxian 触觉手套，仿真复现相同的运动学与相机但不含触觉。
Because Spark-0 uses touch in two ways, its evaluation must answer two
questions: does tactile pretraining improve the visual policy when touch is
absent, and does online touch improve contact-rich control? SparkArena
provides one track for each. Both tracks use the bimanual dexterous Gen 1 Prototype
embodiment, keeping robot morphology and camera layout fixed
(Appendix~\ref{app:sky1}). The real platform carries Moxian tactile gloves
with the same taxel layout as HumanTouch, whereas the simulation reproduces
the kinematics and cameras without touch.

\begin{figure*}[!t]
  \centering
  \includegraphics[width=\textwidth,trim={0 11 0 0},clip]{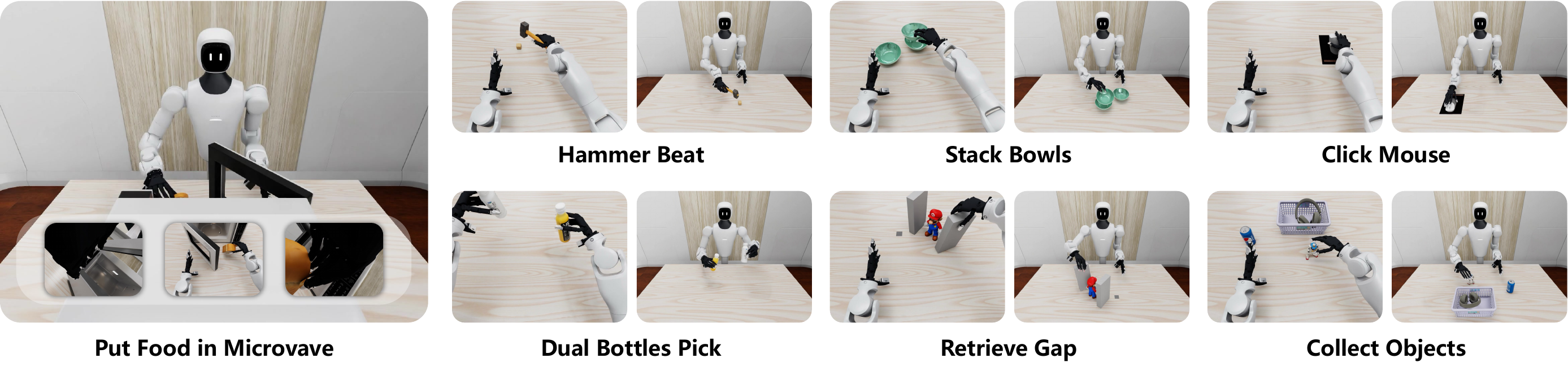}
  % 中文图注：SparkArena-Sim。七项纯视觉灵巧操作仿真任务。
  \caption{\textbf{SparkArena-Sim.} Seven vision-only dexterous tasks in
  simulation.}
  \label{fig:sparkarena-sim}

  \vspace{0.8em}
  \includegraphics[width=\textwidth,trim={0 9 0 0},clip]{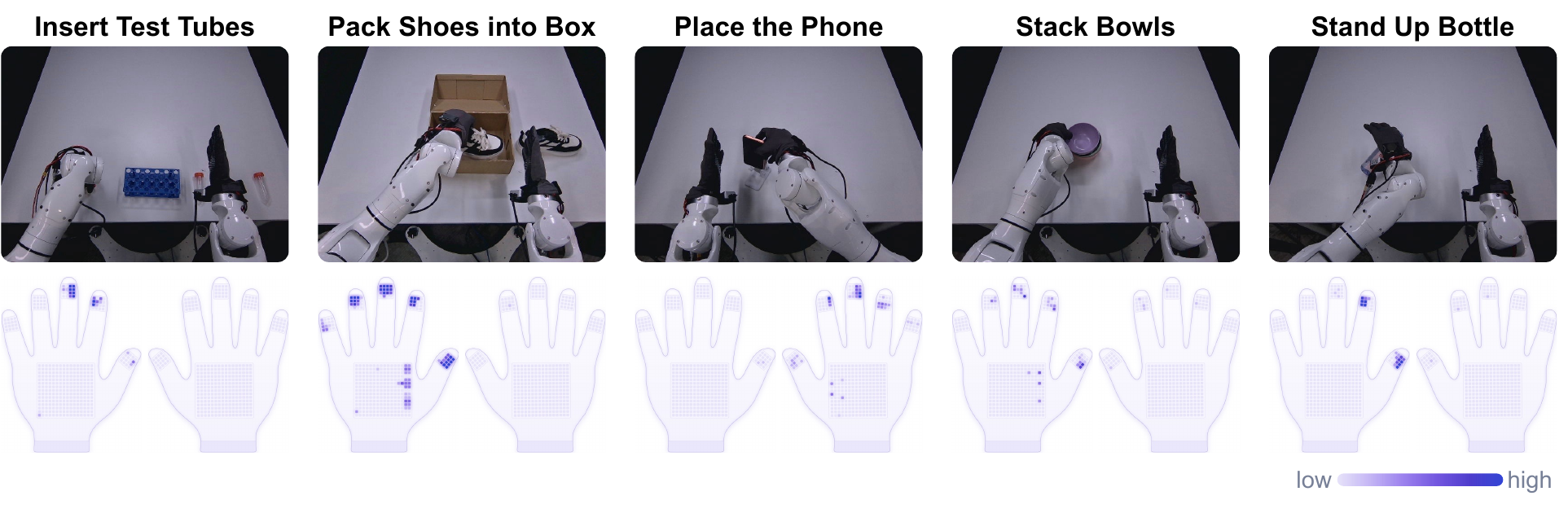}
  % 中文图注：SparkArena-Real。五项接触密集型任务：上为执行画面，下为双手触觉响应。
  \caption{\textbf{SparkArena-Real.} Five contact-rich tasks: rollout views
  (top) and bimanual tactile responses (bottom).}
  \label{fig:sparkarena-real}
\end{figure*}

% 中文：SparkArena-Sim 用七项任务测试纯视觉灵巧控制，覆盖工具使用、精细按压、双手抓取、
% 受限空间取物、堆叠、物体收集和长时序家务操作；仿真使评测可规模化且可复现。
\paragraph{SparkArena-Sim}
This track tests vision-only dexterous control on seven tasks spanning tool
use, precise actuation, bimanual grasping, constrained retrieval, stacking,
object collection, and long-horizon household manipulation
(Figure~\ref{fig:sparkarena-sim}). Simulation keeps evaluation scalable and
reproducible.

% 中文：SparkArena-Real 测试视觉难以观测接触状态时的控制。五项接触密集型任务覆盖插入、
% 柔顺装箱、精确放置、堆叠和物体重定向。真机评测噪声较大，因此所有策略在成对初始条件下
% 评测。我们开源物体、任务定义、成对初始条件和评分协议。
\paragraph{SparkArena-Real}
This track tests control when contact state is poorly observable from
vision. Its five contact-rich tasks span insertion, compliant packing,
precise placement, stacking, and object reorientation
(Figure~\ref{fig:sparkarena-real}). Because real-robot trials are noisy, all
policies are evaluated under paired initial conditions. We release the
objects, task definitions, paired initial conditions, and scoring protocols
(Appendix~\ref{app:sparkarena-real}).

% 中文：评测指标。两条轨道均报告成功率与完成分数。只有完成任务全部目标才算成功；完成分数
% 按子目标进度给分，区分接近成功与早期失败。逐任务判定规则见附录。
\paragraph{Metrics}
Both tracks report success rate and completion score. A trial succeeds only
if it achieves every goal of the task, whereas the completion score credits
progress through subgoals, distinguishing near-successes from early
failures. Per-task criteria are given in
Appendix~\ref{app:sparkarena-scoring}.

\section{Experiments}
\label{sec:experiments}

% 中文：实验研究三个问题：(1) 部署时无触觉，Spark-0 表现如何？(2) 使用在线触觉时，它在
% 接触密集型任务上表现如何？(3) 预训练与 HumanTouch 数据量如何影响各设置？
We study three questions: (1) how Spark-0 performs when touch is unavailable
at deployment, (2) how it performs on contact-rich tasks with online touch,
and (3) how pretraining and the amount of HumanTouch data affect each
setting.

\subsection{Experimental setup}
\label{sec:exp_setup}

% 中文：训练与变体。Spark-0 在第~\ref{sec:data_mixture}节的数据混合上预训练 400k 步，再在
% 各基准的下游数据上微调（附录~\ref{app:downstream}）。评测任务不出现在预训练中。四个变体
% 共享下游数据与优化预算，仅预训练不同。Scratch 不做预训练，从 Qwen3-VL 初始化直接在下游
% 数据上训练，微调目标与其他变体相同。Visual PT 为纯视觉预训练：使用与 1000H 完全相同的预训练数据，仅舍弃全部触觉
% 信号及由触觉生成的接触对齐目标。100H 与 1000H 分别保留 100 和 1,000 小时 HumanTouch 的
% 触觉。三个预训练变体的 RGB 观测与预训练计算量相同，因此 Scratch→Visual PT 衡量纯视觉
% 预训练的作用，Visual PT→1000H 衡量触觉预训练的作用。除非特别说明，Spark-0 指 1000H。
\paragraph{Training and variants}
Spark-0 is pretrained for 400k steps on the mixture in
Section~\ref{sec:data_mixture} and then fine-tuned on each benchmark's
downstream data (Appendix~\ref{app:downstream}). No evaluated task appears
in pretraining. Four variants share the downstream data and optimization
budget and differ only in pretraining. Scratch skips pretraining and trains
the same architecture on downstream data directly from the Qwen3-VL
initialization, with the same fine-tuning objectives. Visual PT is vision-only pretraining: it uses exactly the
pretraining data of 1000H but discards every tactile signal, together with
the touch-derived contact-alignment objective. 100H and 1000H retain touch
for $100$ and $1{,}000$ hours of HumanTouch. The three pretrained variants
share RGB observations and pretraining compute, so Scratch $\rightarrow$
Visual PT measures visual pretraining and Visual PT $\rightarrow$ 1000H
measures tactile pretraining. Unless noted otherwise, Spark-0 refers to
1000H.

% 中文：评测。Spark-0 在 SparkArena-Sim 与 Ego Humanoid 上纯视觉部署，在 SparkArena-Real
% 上使用触觉。SparkArena 报告成功率与完成分数，Ego Humanoid 报告成功率与部分成功率（PSR）。
% 每个策略在 SparkArena-Real 上每任务评测 10 次，SparkArena-Sim 每任务 50 次，Ego Humanoid
% 每任务 seen 27 次、unseen 66 次。
\paragraph{Evaluation}
Spark-0 is deployed without touch on SparkArena-Sim and the Ego Humanoid
Manipulation Benchmark (Ego Humanoid)~\cite{yang2025egovla}, and with touch on SparkArena-Real. We
report success rate together with completion score on SparkArena
(Appendix~\ref{app:sparkarena-scoring}) and partial success rate (PSR) on
Ego Humanoid. Each policy is evaluated for $10$ trials per task on
SparkArena-Real, $50$ on SparkArena-Sim, and $27$ (seen) and $66$ (unseen)
on Ego Humanoid.

% 中文：基线覆盖通用机器人策略、人类先验迁移、世界-动作模型、任务专用模仿、多流控制与
% 触觉基础策略。
\paragraph{Baselines}
We compare against generalist robot policies
($\pi_{0.5}$~\cite{intelligence2025pi_}, GR00T-N1.7~\cite{bjorck2025gr00t},
StarVLA~\cite{community2026starvla}, RDT-1B~\cite{liu2024rdt},
X-VLA~\cite{zheng2025x}), human-prior transfer
(EgoVLA~\cite{yang2025egovla}, H-RDT~\cite{bi2025h},
VITRA~\cite{li2026scalable}, UniT~\cite{chen2026unit}), world-action
modeling (FastWAM~\cite{yuan2026fast}), task-specific imitation
(ACT~\cite{zhao2023learning}), multi-stream control
(RLDX-1~\cite{kim2026rldx}), and tactile foundation policies
(FTP-1~\cite{yuan2026ftp});
details are in Appendix~\ref{app:baselines}.

\subsection{Overall performance}
\label{sec:exp_overall}

\begin{table*}[t]
\centering
% 中文图注：SparkArena-Sim 结果（成功率 % / 完成分数），Spark-0 无触觉输入。全文
% 蓝色为最优，加粗为次优。
\caption{\textbf{SparkArena-Sim results} (SR \% / completion score).
Spark-0 (1000H) runs without touch. Throughout, blue marks the best and bold
the second best.}
\label{tab:spark-arena}
\begingroup
\footnotesize
\setlength{\tabcolsep}{3.5pt}
\renewcommand{\arraystretch}{1.12}
\fitwidth{%
\begin{tabular}{l *{8}{c}}
\toprule
\textbf{Method}
& \makecell{\textbf{Click}\\\textbf{Mouse}}
& \makecell{\textbf{Collect}\\\textbf{Objects}}
& \makecell{\textbf{Dual Bottles}\\\textbf{Pick}}
& \makecell{\textbf{Hammer}\\\textbf{Beat}}
& \makecell{\textbf{Put Food in}\\\textbf{Microwave}}
& \makecell{\textbf{Retrieve}\\\textbf{Gap}}
& \makecell{\textbf{Stack}\\\textbf{Bowls}}
& \textbf{Mean} \\
\midrule
FastWAM~\cite{yuan2026fast}
& 28.00 / 28.00
& \runnerup{8.00} / \runnerup{24.80}
& \runnerup{40.00} / 52.60
& \sota{58.00} / \sota{58.00}
& 32.00 / 32.00
& 2.00 / 2.00
& 30.00 / 37.20
& \runnerup{28.29} / \runnerup{33.51} \\

VITRA~\cite{li2026scalable}
& 14.00 / 14.00
& 6.00 / 17.50
& \sota{62.00} / \sota{71.00}
& 46.00 / 46.00
& 32.00 / 32.00
& 2.00 / 2.00
& 24.00 / 28.80
& 26.57 / 30.19 \\

StarVLA~\cite{community2026starvla}
& 16.00 / 16.00
& \runnerup{8.00} / 17.80
& 36.00 / 48.60
& 50.00 / 50.00
& 44.00 / 44.00
& \runnerup{8.00} / \runnerup{8.00}
& 22.00 / 27.70
& 26.29 / 30.30 \\

$\pi_{0.5}$~\cite{intelligence2025pi_}
& 14.00 / 14.00
& \sota{10.00} / \sota{26.60}
& 32.00 / 50.00
& 48.00 / 48.00
& 38.00 / 38.00
& 2.00 / 2.00
& \sota{38.00} / \sota{43.70}
& 26.00 / 31.76 \\

H-RDT~\cite{bi2025h}
& \runnerup{30.00} / \runnerup{30.00}
& 6.00 / 13.20
& 28.00 / 39.40
& \runnerup{52.00} / \runnerup{52.00}
& \runnerup{48.00} / \runnerup{48.00}
& 0.00 / 0.00
& 10.00 / 12.40
& 24.86 / 27.86 \\

UniT~\cite{chen2026unit}
& 10.00 / 10.00
& 2.00 / 10.20
& 14.00 / 27.20
& 42.00 / 42.00
& 40.00 / 40.00
& 0.00 / 0.00
& 8.00 / 15.20
& 16.57 / 20.66 \\

ACT~\cite{zhao2023learning}
& 22.00 / 22.00
& 0.00 / 5.90
& 2.00 / 9.20
& 22.00 / 22.00
& \sota{52.00} / \sota{52.00}
& 0.00 / 0.00
& 2.00 / 4.70
& 14.29 / 16.54 \\

GR00T-N1.7~\cite{bjorck2025gr00t}
& 6.00 / 6.00
& 2.00 / 10.70
& 34.00 / 47.80
& 34.00 / 34.00
& 20.00 / 20.00
& 0.00 / 0.00
& 2.00 / 7.40
& 14.00 / 17.99 \\

RLDX-1~\cite{kim2026rldx}
& 4.00 / 4.00
& 0.00 / 13.90
& 28.00 / 43.00
& 26.00 / 26.00
& 18.00 / 18.00
& 0.00 / 0.00
& 10.00 / 11.80
& 12.29 / 16.67 \\

RDT-1B~\cite{liu2024rdt}
& 4.00 / 4.00
& 0.00 / 6.10
& 6.00 / 16.20
& 26.00 / 26.00
& 32.00 / 32.00
& 0.00 / 0.00
& 0.00 / 3.30
& 9.71 / 12.51 \\

EgoVLA~\cite{yang2025egovla}
& 0.00 / 0.00
& 0.00 / 4.60
& 8.00 / 14.60
& 14.00 / 14.00
& 16.00 / 16.00
& 0.00 / 0.00
& 0.00 / 1.20
& 5.43 / 7.20 \\

\midrule

Spark-0 (1000H)
& \sota{42.00} / \sota{42.00}
& \runnerup{8.00} / 14.10
& \sota{62.00} / \runnerup{70.40}
& \runnerup{52.00} / \runnerup{52.00}
& \sota{52.00} / \sota{52.00}
& \sota{10.00} / \sota{10.00}
& \runnerup{32.00} / \runnerup{37.70}
& \sota{36.86} / \sota{39.74} \\
\bottomrule
\end{tabular}}
\endgroup
\end{table*}

\begin{table*}[t]
\centering
% 中文图注：Ego Humanoid 结果（平均 SR / PSR），Spark-0 纯视觉部署；逐任务结果见附录。
\caption{\textbf{Ego Humanoid results} (mean SR / partial success rate PSR).
Spark-0 (1000H) runs without touch; per-task results are in
Appendix~\ref{app:egovla_detailed}.}
\label{tab:egovla-summary}
\begingroup
\scriptsize
\setlength{\tabcolsep}{3.5pt}
\renewcommand{\arraystretch}{1.12}
\fitwidth{%
\begin{tabular}{l *{8}{c}}
\toprule
& \multicolumn{4}{c}{Short-Horizon Benchmark}
& \multicolumn{4}{c}{Long-Horizon Benchmark} \\
\cmidrule(lr){2-5}\cmidrule(lr){6-9}
Method
& \multicolumn{2}{c}{Seen}
& \multicolumn{2}{c}{Unseen}
& \multicolumn{2}{c}{Seen}
& \multicolumn{2}{c}{Unseen} \\
\cmidrule(lr){2-3}\cmidrule(lr){4-5}
\cmidrule(lr){6-7}\cmidrule(lr){8-9}
& SR $\uparrow$ & PSR $\uparrow$
& SR $\uparrow$ & PSR $\uparrow$
& SR $\uparrow$ & PSR $\uparrow$
& SR $\uparrow$ & PSR $\uparrow$ \\
\midrule

EgoVLA~\cite{yang2025egovla}
& \runnerup{77.78}
& \runnerup{84.92}
& \runnerup{69.11}
& \runnerup{76.26}
& 45.93
& \runnerup{80.78}
& 28.79
& 69.11 \\

FastWAM~\cite{yuan2026fast}
& 72.49
& 83.25
& 32.47
& 45.42
& \runnerup{49.63}
& 78.00
& 21.21
& 39.27 \\

UniT~\cite{chen2026unit}
& 67.72
& 75.75
& 61.69
& 70.92
& 5.19
& 44.59
& 6.36
& 46.06 \\

VITRA~\cite{li2026scalable}
& 63.49
& 76.46
& 59.31
& 74.82
& 40.00
& 71.41
& \runnerup{39.70}
& \runnerup{74.18} \\

RLDX-1~\cite{kim2026rldx}
& 57.67
& 71.96
& 53.68
& 68.36
& 6.67
& 40.44
& 6.06
& 39.82 \\

GR00T-N1.7~\cite{bjorck2025gr00t}
& 55.56
& 70.46
& 55.63
& 69.05
& 0.74
& 37.19
& 1.82
& 40.58 \\

H-RDT~\cite{bi2025h}
& 51.85
& 67.11
& 34.85
& 53.21
& 14.81
& 54.26
& 9.09
& 46.18 \\

$\pi_{0.5}$~\cite{intelligence2025pi_}
& 46.56
& 62.70
& 34.20
& 51.01
& 19.26
& 49.59
& 16.06
& 46.29 \\

StarVLA~\cite{community2026starvla}
& 43.39
& 54.76
& 40.48
& 51.77
& 17.04
& 52.67
& 16.97
& 51.80 \\

RDT-1B~\cite{liu2024rdt}
& 32.28
& 51.76
& 34.20
& 51.26
& 10.37
& 51.30
& 10.91
& 52.55 \\

\midrule

Spark-0 (1000H)
& \sota{80.95}
& \sota{88.18}
& \sota{75.97}
& \sota{83.80}
& \sota{68.15}
& \sota{80.96}
& \sota{61.82}
& \sota{80.53} \\

\bottomrule
\end{tabular}}
\endgroup
\end{table*}
\begin{table*}[t]
\centering
% 中文图注：SparkArena-Real 结果（成功率 % / 完成分数），Spark-0 使用触觉；全部变体见
% 表~\ref{tab:real-bench-all}。
\caption{\textbf{SparkArena-Real results} (SR \% / completion score).
Spark-0 (1000H) uses touch; all variants are in
Table~\ref{tab:real-bench-all}.}
\label{tab:real-bench}
\begingroup
\footnotesize
\setlength{\tabcolsep}{3.5pt}
\renewcommand{\arraystretch}{1.12}
\fitwidth{%
\begin{tabular}{l *{6}{c}}
\toprule
\textbf{Method}
& \makecell{\textbf{Insert}\\\textbf{Test Tubes}}
& \makecell{\textbf{Pack Shoes}\\\textbf{into Box}}
& \makecell{\textbf{Place}\\\textbf{the Phone}}
& \makecell{\textbf{Stack}\\\textbf{Bowls}}
& \makecell{\textbf{Stand Up}\\\textbf{Bottle}}
& \textbf{Mean} \\
\midrule
H-RDT~\cite{bi2025h}
& 0.00 / 0.00 & 0.00 / 0.00 & 0.00 / 0.00
& 0.00 / 0.00 & 0.00 / 0.00 & 0.00 / 0.00 \\

RDT-1B~\cite{liu2024rdt}
& 0.00 / 0.00 & 0.00 / 0.00 & 0.00 / 0.00
& 0.00 / 0.00 & 0.00 / 0.00 & 0.00 / 0.00 \\

$\pi_{0.5}$~\cite{intelligence2025pi_}
& 0.00 / 8.00 & 10.00 / 50.00 & 0.00 / 15.00
& \sota{50.00} / \sota{73.00} & 10.00 / 10.00
& 14.00 / 31.20 \\

GR00T-N1.7~\cite{bjorck2025gr00t}
& 0.00 / 7.00 & 10.00 / 30.00 & 0.00 / 12.00
& 20.00 / 51.00 & 50.00 / 50.00 & 16.00 / 30.00 \\

X-VLA~\cite{zheng2025x}
& 0.00 / 17.00 & 10.00 / 10.00
& 10.00 / 10.00 & \runnerup{30.00} / 54.00
& \runnerup{60.00} / \runnerup{60.00}
& 22.00 / 30.20 \\

StarVLA~\cite{community2026starvla}
& 0.00 / 0.00 & \runnerup{30.00} / 55.00
& 10.00 / 10.00 & 20.00 / 43.00
& 30.00 / 30.00 & 18.00 / 27.60 \\

FTP-1~\cite{yuan2026ftp}
& 0.00 / \runnerup{24.00} & \sota{80.00} / \runnerup{80.00}
& \runnerup{20.00} / \runnerup{32.00} & \runnerup{30.00} / 53.00
& 50.00 / 50.00 & \runnerup{36.00} / \runnerup{47.80} \\

\midrule

Spark-0 (1000H)
& 0.00 / \sota{31.00}
& \sota{80.00} / \sota{85.00}
& \sota{50.00} / \sota{56.00}
& \sota{50.00} / \runnerup{62.00}
& \sota{80.00} / \sota{80.00}
& \sota{52.00} / \sota{62.80} \\
\bottomrule
\end{tabular}}
\endgroup
\end{table*}

% 中文：目标本体上的纯视觉控制。Spark-0 在 SparkArena-Sim 上达到 36.86% 成功率与 39.74
% 完成分数，较最强基线 FastWAM 高 8.57 和 6.23，并在七项任务中的四项取得或并列最高成功率。
% 各基线只擅长个别技能，而 Spark-0 在精细与双手任务上均保持竞争力。GR00T-N1.7 与
% RDT-1B 尽管经过大规模多机器人预训练，仅为 14.00% 和 9.71%，表明广泛的机器人预训练本身
% 不足以支撑高自由度灵巧控制。
\paragraph{Vision-only control on the target embodiment}
On SparkArena-Sim (Table~\ref{tab:spark-arena}), Spark-0 reaches $36.86\%$
success and a $39.74$ completion score, exceeding the strongest baseline,
FastWAM, by $8.57$ and $6.23$ points, and achieves or ties the best success
rate on four of seven tasks. Baselines excel only on individual skills
(FastWAM on Hammer Beat, VITRA on Dual Bottles Pick, $\pi_{0.5}$ on Collect
Objects and Stack Bowls), whereas Spark-0 remains competitive across
precise and bimanual tasks. GR00T-N1.7 and RDT-1B reach only
$14.00\%$ and $9.71\%$ despite large-scale multi-robot pretraining,
suggesting that broad robot pretraining alone is insufficient for high-DoF
dexterous control.

% 中文：跨本体泛化。Spark-0 在 Ego Humanoid 全部八项汇总指标上排名第一，且优势随任务时长
% 扩大：相对最强基线的成功率优势从短时序 seen/unseen 的 3.17/6.86 增至长时序的
% 18.52/22.12。在 seen 长时序任务上，其部分成功率与最佳基线持平（+0.18），成功率却高出
% 18.52，说明优势在于完成完整子目标序列。基线优势分散：EgoVLA 是最强短时序基线，但在 unseen
% 长时序任务上降至 28.79%；通用机器人 VLA 的 seen 长时序成功率均低于 20%。评测不使用触觉，
% 因此结果仅反映视觉策略。
\paragraph{Cross-embodiment generalization}
Spark-0 ranks first on all eight Ego Humanoid aggregate metrics
(Table~\ref{tab:egovla-summary}), and its margin grows with task horizon:
its success advantage over the strongest baseline rises from $3.17$/$6.86$
points on seen/unseen short-horizon tasks to $18.52$/$22.12$ points on
long-horizon tasks. On seen long-horizon tasks, Spark-0 matches the best
baseline in partial success ($+0.18$) but leads by $18.52$ points in
success: its advantage lies in completing full subgoal sequences. Baseline strengths are fragmented: EgoVLA is the strongest
short-horizon baseline but falls to $28.79\%$ on unseen long-horizon tasks,
and generalist robot VLAs remain below $20\%$ on seen long-horizon success.
Since touch is absent at evaluation, these results reflect the visual policy
alone.

% 中文：接触密集型真机控制。Spark-0 在 SparkArena-Real 上达到 52.00% 成功率与 62.80 完成
% 分数，较最强基线——触觉基础策略 FTP-1（36.00%，47.80）——高 16.00 和 15.00，并在五项任务中
% 的四项取得最高完成分数。纯视觉基线常能启动正确行为，但难以完成接触序列：π0.5 和 GR00T-N1.7
% 完成分数约 30，成功率仅 14.00% 和 16.00%；RDT-1B 与 H-RDT 全部失败。Spark-0 相对 FTP-1
% 的优势在 Place the Phone（50.00% 对 20.00%）和
% Stand Up Bottle（80.00% 对 50.00%）上最为明显。Insert Test Tubes 需插满三支试管，没有方法
% 完全成功，但 Spark-0 取得最高完成分数（31.00），体现更好的精细插入进度。
\paragraph{Contact-rich real-robot control}
On SparkArena-Real (Table~\ref{tab:real-bench}), Spark-0 reaches $52.00\%$
success and a $62.80$ completion score, exceeding the strongest baseline,
the tactile foundation policy FTP-1 ($36.00\%$ and $47.80$), by $16.00$ and
$15.00$ points, with the highest completion score on four of five tasks.
Vision-only baselines often initiate the correct behavior but rarely
complete the contact sequence: $\pi_{0.5}$ and GR00T-N1.7 reach completion
scores near $30$ but only $14.00\%$ and $16.00\%$ success, and RDT-1B and
H-RDT fail every task. Spark-0 widens its margin over FTP-1 most clearly
on Place the Phone ($50.00\%$ vs.\ $20.00\%$) and Stand Up Bottle ($80.00\%$ vs.\
$50.00\%$). Insert Test Tubes, which requires filling all three slots, is
not fully completed by any method, yet Spark-0 attains the highest
completion score ($31.00$), reflecting finer insertion progress. Per-task
results are in
Appendices~\ref{app:egovla_detailed} and~\ref{app:real_all}.

\subsection{Pretraining and tactile-scale analysis}
\label{sec:exp_ablations}

\begin{table*}[t]
\centering
% 中文图注：Spark-0 消融（平均 SR / 分数）。Scratch 不预训练；Visual PT 使用与 1000H 相同的
% 数据但舍弃触觉；100H/1000H 保留 100/1,000 小时 HumanTouch 触觉。分数在 SparkArena 为完成分数，
% 在 Ego Humanoid 为 PSR（四个设置平均）。
\caption{\textbf{Spark-0 pretraining comparison} (mean SR / score). Scratch
is not pretrained; Visual PT uses the 1000H pretraining data with touch
discarded; 100H/1000H retain touch for 100/$1{,}000$ hours of HumanTouch. Score is
completion on SparkArena and PSR on Ego Humanoid, averaged over the four Ego
Humanoid settings.}
\label{tab:spark0-ablation}
\begingroup
\small
\setlength{\tabcolsep}{10pt}
\renewcommand{\arraystretch}{1.12}
\begin{tabular}{l *{3}{c}}
\toprule
Configuration
& \makecell{SparkArena-Sim\\SR / Score}
& \makecell{Ego Humanoid\\SR / Score}
& \makecell{SparkArena-Real\\SR / Score} \\
\midrule
Spark-0 (Scratch)
& 31.71 / 35.17
& 58.34 / 73.71
& 38.00 / 51.80 \\

Spark-0 (Visual PT)
& \runnerup{36.57} / \sota{39.84}
& 52.45 / 69.89
& \runnerup{44.00} / \runnerup{61.80} \\

Spark-0 (100H)
& 35.14 / 38.01
& \runnerup{59.42} / \runnerup{76.72}
& 40.00 / 52.00 \\
\midrule
Spark-0 (1000H)
& \sota{36.86} / \runnerup{39.74}
& \sota{71.72} / \sota{83.37}
& \sota{52.00} / \sota{62.80} \\
\bottomrule
\end{tabular}
\endgroup
\end{table*}

% 中文：预训练与 HumanTouch。Scratch 本身已是强基础：在两条 SparkArena 轨道上均超过最强基线（Sim 31.71% 对 28.29%，Real 38.00% 对 36.00%），预训练在此之上带来进一步提升。纯视觉预训练
% （Scratch→Visual PT）提升 SparkArena-Sim（+4.86）与 SparkArena-Real（+6.00），但使
% Ego Humanoid 从 58.34% 降至 52.45%。在相同 RGB 上加入 1,000 小时触觉（Visual PT→1000H），
% SparkArena-Sim 基本不变（+0.29），Ego Humanoid 提升 19.27。由于两者 RGB 数据完全相同，该收益
% 由触觉驱动，经由视觉 token 的接触图监督与预训练的触觉通路实现；Ego Humanoid 为纯视觉评测，
% 因此收益来自触觉预训练而非在线触觉反馈。一种解释是接触监督改善了向未见视觉域的迁移，而 Gen 1 Prototype
% 数据已出现在预训练中，因此该作用在 SparkArena-Sim 上不明显。在 SparkArena-Real 上，所有变体
% 均以触觉微调和部署，1000H 较 Visual PT 将成功率从 44.00% 提升至 52.00%，体现触觉预训练的贡献。
\paragraph{Pretraining and HumanTouch}
Table~\ref{tab:spark0-ablation} compares the four variants. Scratch is
already a strong foundation, exceeding the strongest baseline on both
SparkArena tracks ($31.71\%$ vs.\ $28.29\%$ on Sim and $38.00\%$ vs.\
$36.00\%$ on Real), and pretraining builds further on it. Vision-only
pretraining (Scratch $\rightarrow$ Visual PT) improves SparkArena-Sim
($+4.86$) and SparkArena-Real ($+6.00$) but lowers Ego Humanoid from
$58.34\%$ to $52.45\%$. Adding $1{,}000$ hours of touch on the same RGB data
(Visual PT $\rightarrow$ 1000H) leaves SparkArena-Sim nearly unchanged
($+0.29$) but raises Ego Humanoid by $19.27$ points. Because the two
variants share identical RGB data, this gain is driven by touch, acting
through contact-map supervision of the visual tokens and the pretrained
tactile pathway; because Ego Humanoid is evaluated without touch, it comes
from tactile pretraining rather than online tactile feedback. One
explanation is that contact supervision improves transfer to an unseen
visual domain, an effect that matters little on SparkArena-Sim because
Gen 1 Prototype data already appear in pretraining. On SparkArena-Real, every variant
is fine-tuned and deployed with touch, and 1000H raises success from
$44.00\%$ to $52.00\%$ over Visual PT, reflecting the contribution of
tactile pretraining.

% 中文：触觉规模。在 Ego Humanoid 上，成功率随触觉时长单调提升（Visual PT 52.45%、100H
% 59.42%、1000H 71.72%）。在 SparkArena-Sim 与 SparkArena-Real 上，100H 未超过 Visual PT，
% 1000H 最佳。触觉预训练的收益在完整的 1000 小时规模下最为明显：只有该配置在三项基准上均
% 取得最佳成功率。
\paragraph{Tactile scale}
On Ego Humanoid, success rises with the hours of touch, from $52.45\%$
(Visual PT) to $59.42\%$ (100H) and $71.72\%$ (1000H). On SparkArena-Sim
and SparkArena-Real, 100H does not exceed Visual PT, while 1000H performs
best. The benefit of tactile pretraining is thus most evident at the full
1000-hour scale, the only configuration that achieves the best success rate
on all three benchmarks.

% 中文：主要发现。综合来看，实验得出三点发现。其一，部署时无触觉，Spark-0 在目标本体和
% 未见人形本体上均优于全部基线，且在 Ego Humanoid 上优势随任务时长扩大。其二，使用在线触觉时，
% 它在接触密集型真机任务上的成功率比最强基线（同为触觉策略）高 16.00 个百分点。其三，跨本体
% 收益来自触觉预训练，并在完整规模下最为明显：1,000 小时 HumanTouch 使 Ego Humanoid 纯视觉
% 成功率提升 19.27 个百分点，且只有该配置在三项基准上均取得最佳成功率。
\paragraph{Key findings}
Taken together, the experiments yield three findings. First, without touch
at deployment, Spark-0 outperforms all baselines on both the target
embodiment and an unseen humanoid, and on Ego Humanoid its margin grows
with task horizon. Second, with online touch, it exceeds the strongest
baseline, itself a tactile policy, by $16.00$ points in success on
contact-rich real-robot tasks. Third, the cross-embodiment gain stems from
tactile pretraining and is most evident at full scale: $1{,}000$ hours of
HumanTouch raise vision-only success on Ego Humanoid by $19.27$ points, and
only this configuration achieves the best success rate on all three
benchmarks.

\section{Conclusion}

% 中文：我们提出空间对齐的 VTLA 模型 Spark-0，以及 1,000 小时全手触觉数据集 HumanTouch
% 和配对纯视觉仿真任务与触觉条件化真机任务的基准 SparkArena。Spark-0 将触觉用作视觉 token
% 的仅训练监督和动作专家的输入，单一 checkpoint 同时支持纯视觉与触觉条件化部署。它在两条
% SparkArena 轨道上取得最高平均成功率，并在 Ego Humanoid 全部汇总指标上排名第一。1,000 小时
% HumanTouch 触觉预训练在未见的 Ego Humanoid 本体上、纯视觉部署时带来最大收益，表明触觉不仅是
% 策略输入，也是灵巧视觉运动预训练的可规模化监督信号。
We presented Spark-0, a spatially aligned VTLA model, together with
HumanTouch, a $1{,}000$-hour full-hand tactile dataset, and SparkArena, a
benchmark that pairs vision-only simulated tasks with tactile-conditioned
real-robot tasks. Spark-0 uses touch both as train-only supervision of
visual tokens and as an input to the action expert, so a single checkpoint
supports vision-only and tactile-conditioned deployment. It achieves the
highest mean success rate on both SparkArena tracks and ranks first on all
Ego Humanoid aggregate metrics. Tactile pretraining on $1{,}000$ hours of
HumanTouch yields its largest gain on the unseen Ego Humanoid embodiment
under vision-only deployment, indicating that touch is not only a policy input but also scalable supervision for
dexterous visuomotor pretraining.

\bibliography{reference}

\clearpage
\appendix
\begin{center}
  {\Large\bfseries Appendix}
\end{center}
\vspace{0.5em}

\section{HumanTouch Dataset Details}
\label{app:humantouch_details}

\begin{figure}[!htb]
  \centering
  \includegraphics[width=0.85\linewidth,trim={0 3.5 0 0},clip]{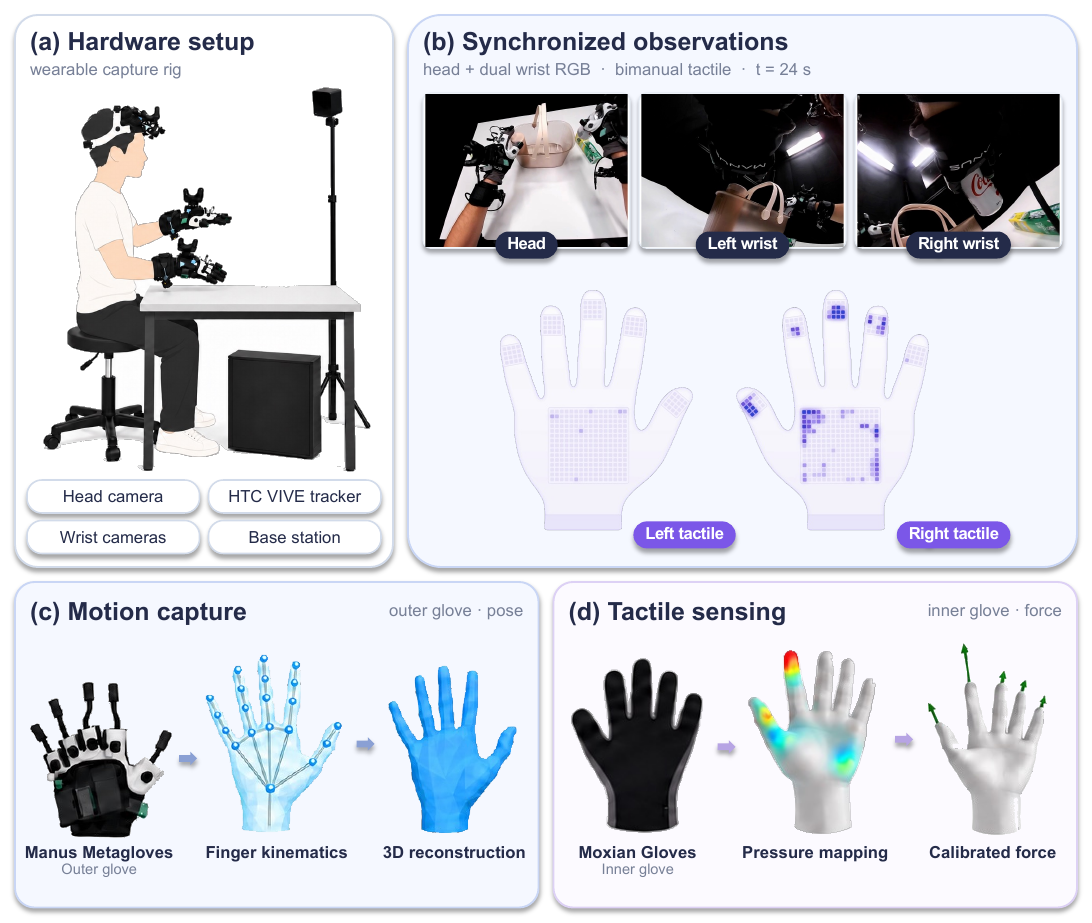}
  % 中文图注：HumanTouch 采集硬件。可穿戴装置同步多视角 RGB、手部动捕与双手触觉手套。
  \caption{\textbf{HumanTouch acquisition hardware.} Multi-view RGB, hand
  motion capture, and bilateral tactile gloves are synchronized on a
  wearable rig.}
  \label{fig:humantouch-hardware}
\end{figure}

\begin{figure}[!htb]
  \centering
  \includegraphics[width=0.38\linewidth,trim={0 24 0 0},clip]{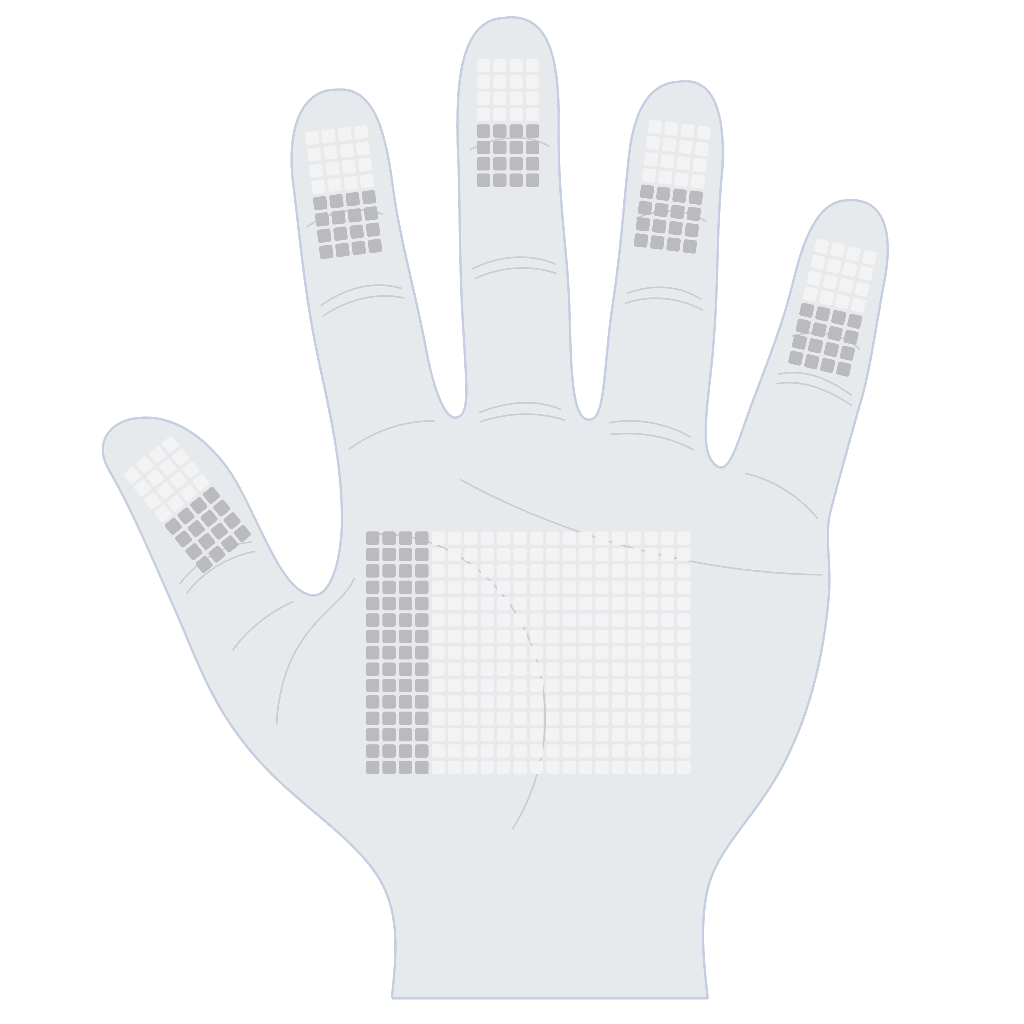}
  % 中文图注：手套感知点布局（右手掌侧视图）。每只手共 460 个感知点；深色的 140 个位于
  % 弯折区域，易因弯曲产生伪接触压力而被屏蔽，浅色的 320 个用于模型观测。
  \caption{\textbf{Glove taxel layout} (right hand, palm view). Each glove has
  $460$ sensing points per hand. The $140$ dark points over flexion regions
  are masked to avoid bending-induced spurious pressure; the $320$ light
  points form the model observation.}
  \label{fig:glove-taxel-mask}
\end{figure}

% 中文：HumanTouch 包含 1,000 小时操作数据，覆盖 100 余项任务和 32 名操作者，在可穿戴
% 设备上以 60 Hz 同步记录头部与腕部视频、双手全手触觉、手指关节追踪和手腕位姿
% （图~\ref{fig:humantouch-hardware}）。每只手套有 460 个感知点（图~\ref{fig:glove-taxel-mask}）；
% 位于弯折区域的 140 个（每根手指靠近掌侧的 16 个，以及手掌拇指侧 4 列共 60 个）因弯曲会
% 产生伪压力而被屏蔽。其余 320 个 taxel 构成模型观测：taxel 0--239 为手掌区域，其后每 16
% 个为一个手指区域，每只手共六个解剖区域。触觉编码器（附录~\ref{app:tactile_encoder}）与
% 接触图生成（附录~\ref{app:contact_maps}）均使用这一布局。
HumanTouch contains $1{,}000$ hours of manipulation from more than $100$
tasks and $32$ operators. The wearable rig records synchronized head- and
wrist-camera video, bilateral full-hand tactile signals, finger-joint
tracking, and wrist poses at $60$~Hz
(Figure~\ref{fig:humantouch-hardware}). Each glove carries $460$ sensing
points per hand (Figure~\ref{fig:glove-taxel-mask}). The $140$ points over
flexion regions, namely the $16$ proximal points of each finger and the $60$
points in the four thumb-side palm columns, are masked because bending there
induces spurious pressure. The remaining $320$ taxels form the model
observation: taxels $0$--$239$ constitute the palm region and each
subsequent 16-taxel block constitutes one finger region, giving six
anatomical regions per hand. The tactile encoder
(Appendix~\ref{app:tactile_encoder}) and contact-map generation
(Appendix~\ref{app:contact_maps}) both use this layout.

\section{Architecture Details}
\label{app:architecture_details}

\subsection{Action expert}

% 中文：动作专家的 block 数 L 与 VLM 层数相同。VLM 隐状态从 1 计数，DiT block 从 0 计数：
% 偶数 block $i$ 交叉注意投影后的 VLM 隐状态 $h^{(i+1)}$，奇数 block 在 State--Tactile--Action
% 联合序列上执行带掩码的自注意力（图~\ref{fig:spark0_architecture}(b)）。每个 block 只执行
% 一次注意力操作，其后接前馈子层。Action token 使用可学习位置嵌入；触觉 token 由流、手和
% 区域嵌入区分，不平移预训练的动作位置。每个 block 的注意力前归一化由共享条件 $c$ 做 AdaLN
% 调制，其中力窗口经零初始化 MLP（隐藏宽度 512）输入，且独立于触觉 token 门控；无手套样本的
% 力项为零。视觉 token 剪枝与可选的未来接触预测头均关闭。
The action expert has as many blocks $L$ as the VLM has layers. VLM hidden
states are indexed from $1$ and DiT blocks from $0$: even block $i$
cross-attends to the projected VLM hidden state $h^{(i+1)}$, and odd blocks
run masked self-attention over the joint State--Tactile--Action sequence
(Figure~\ref{fig:spark0_architecture}(b)). Each block performs exactly one
attention operation followed by a feed-forward sublayer. Action tokens
receive learned positional embeddings, whereas tactile tokens are
distinguished by stream, hand, and region embeddings and therefore do not
shift the pretrained action positions. Pre-attention normalization in every
block is AdaLN-modulated by the shared condition
\begin{equation}
  c
  =
  \operatorname{Embed}(\tau)
  +
  \operatorname{Embed}\!\left(
    \operatorname{vec}(\psi(F_{\mathrm{left}}),\psi(F_{\mathrm{right}}))
  \right),
\end{equation}
where $\psi$ is the pressure transform of
Appendix~\ref{app:tactile_encoder}. The second term flattens the
$2\times6\times20=240$-dimensional regional force window and passes it
through a zero-initialized MLP with hidden width $512$, independent of the
tactile token gates; it is zero for samples without a glove. Visual token
pruning and the optional future-contact head are disabled.

\subsection{Tactile encoder}
\label{app:tactile_encoder}

% 中文：两路编码。区域流对每只手六个区域、20 帧（60 Hz）的力信号使用三层时序卷积（核大小
% 3，隐藏宽度 512）并在时间维平均池化，每个区域产生一个 token。3D 流只使用当前帧：每只手
% 320 个 taxel，各含头部相机坐标系 xyz 与压力；四频段 Fourier 特征经两层、宽度 512 的点 MLP
% 后，在每个区域内做 softmax 注意力池化。两路加入手部与区域嵌入后投影到 1024 维并拼接，
% 不使用 token mixer，共 12 个区域 token 和 12 个 3D token。
\paragraph{Streams}
The regional stream receives, for each hand, six regional force signals
over $20$ frames at $60$~Hz. A three-layer temporal convolution with kernel
size $3$ and hidden width $512$, followed by mean pooling over time, yields
one token per region. The 3D stream receives the current frame only: $320$
taxels per hand, each with head-camera-frame xyz and pressure. Four-band
Fourier features pass through a two-layer, width-512 point MLP, and softmax
attention pools the taxels within each region. For hand $k$, region $r$, and
taxel set $\mathcal{R}_{k,r}$,
\begin{equation}
  e^{\mathrm{reg}}_{k,r}
  =
  f_{\mathrm{temp}}\!\left(\psi(F_{k,r,1:T_F})\right),
  \qquad
  e^{\mathrm{3d}}_{k,r}
  =
  f_{\mathrm{set}}\!\left(
    \operatorname{Pool}_{i\in\mathcal{R}_{k,r}}
    f_{\mathrm{pt}}\!\left(\Gamma([x_i,\psi(p_i)])\right)
  \right).
\end{equation}
After adding learned hand and region embeddings, both streams are projected
to width $1024$ and concatenated without a token mixer, giving $12$
regional and $12$ 3D tokens.

% 中文：压力变换与质量检查。压力经 $\psi(p)=\log(1+p)/s$ 变换，尺度 $s$ 自动估计，不做
% 中心化；零压力 taxel 仍视为有效。若某行 3D 数据中有限值的坐标与压力少于 25%，或超过 85%
% 的值在幅值 1.5 以上饱和，则以占位符替换；卡死的 taxel 在池化前移除。训练时 3D 流以 0.15
% 的概率丢弃，区域流不做模态 dropout。
\paragraph{Pressure conditioning and quality checks}
Pressure is transformed as $\psi(p)=\log(1+p)/s$ with an automatically
estimated scale $s$ and no centering; zero-pressure taxels remain valid. A
3D row is replaced by the placeholder if fewer than $25\%$ of its
coordinates and pressures are finite or more than $85\%$ are saturated above
magnitude $1.5$, and stuck taxels are removed before pooling. During
training, the 3D stream is dropped with probability $0.15$, whereas the
regional stream has no modality dropout.

% 中文：门控与缺失流。存在的触觉流由可学习标量门控缩放，初值 0.1，不加 L1 惩罚；缺失的流
% 由单个可学习占位向量填充并复制到各槽位，占位符不经过门控。$m^q$ 表示流 $q$ 是否存在。
\paragraph{Gating and missing streams}
Present streams are scaled by learned scalar gates $\gamma_{\mathrm{reg}}$
and $\gamma_{\mathrm{3d}}$, both initialized to $0.1$ without an L1 penalty.
Absent streams are filled with a single learned placeholder replicated
across their slots and left ungated:
\begin{equation}
  \tilde e^q
  =
  m^q\gamma_q e^q+(1-m^q)e^\varnothing,
  \qquad q\in\{\mathrm{reg},\mathrm{3d}\},
\end{equation}
where $m^q\in\{0,1\}$ indicates whether stream $q$ is present.

\subsection{Joint attention and tactile shortcut}

% 中文：非对称偏置。联合自注意力中，触觉 query 读取 State 与 Action key 时加入可学习加性
% 偏置 $b(g)$，$g$ 初始化为 1.0；State 与 Action query 读取触觉 key 时不加偏置，该方向的
% 触觉影响只由值门控调节。
\paragraph{Asymmetric bias}
In joint self-attention, tactile queries attending to State and Action keys
receive the additive bias
\begin{equation}
  b(g)=-6\bigl(1-\sigma(g)\bigr),
\end{equation}
with $g$ learned and initialized to $1.0$. State and Action queries attend to
tactile keys without bias, so tactile influence in this direction is
controlled only by the value gates.

% 中文：触觉捷径。训练时，对含触觉的样本，捷径将所有偶数 block 中 3 个 State 行和 50 个
% Action 行的交叉注意力输出置零，而 24 个触觉行仍读取 VLM，使视觉-语言上下文只能经由触觉行
% 到达 Action。其概率在前 20k 步内从 0 线性升至 0.1（微调时为 8k 步），推理时关闭。
\paragraph{Tactile shortcut}
During training on samples with touch, the shortcut zeros the
cross-attention outputs of the $3$ State and $50$ Action rows in every even
block, while the $24$ tactile rows continue to read the VLM, so that
vision-language context reaches the Action rows only through the tactile
rows. Its probability ramps linearly from $0$ to $0.1$ over the first 20k
steps (8k during fine-tuning), and it is disabled at inference.

\section{Train-Only Spatial Supervision}
\label{app:spatial_supervision}

% 中文：token 对应关系。两个目标都作用于 Qwen3-VL 合并后的视觉 token，由图像占位符定位。
% Qwen3-VL 以 16 像素 patch 编码并做 2×2 合并，缩放到 H×W 的视角产生按行优先排列的
% (H/32)×(W/32) 网格：头部 384×384 为 12×12，腕部 256×256 为 8×8。每个目标都重采样到同一
% 网格，使第 j 个目标与第 j 个 token 覆盖相同图像区域。
\paragraph{Token correspondence}
Both objectives act on the merged visual tokens of Qwen3-VL, located by the
image placeholder token. Qwen3-VL encodes $16$-pixel patches and merges them
$2\times2$, so a view resized to $H\times W$ yields an $(H/32)\times(W/32)$
token grid in raster order: $12\times12$ for the $384\times384$ head view
and $8\times8$ for the $256\times256$ wrist views. Every target is resampled
onto this grid, so target $j$ covers the same image cell as token $j$.

% 中文：几何对齐。冻结的 DA3-LARGE 教师接收与 VLM 相同的图像，按 336 像素预算保持宽高比
% 缩放（方形视角为 24×24 个 14 像素 patch），输出最后一层（索引 23）的 1024 维 patch 特征，
% 加上缩放 0.1 的二维正弦位置编码后双线性插值到 token 网格。VLM 第 25 层 token 经
% LayerNorm--MLP 投影器（隐藏宽度 2048）映射到 1024 维。每个视角以 bfloat16 独立处理。
\paragraph{Geometry alignment}
A frozen DA3-LARGE teacher receives the same image as the VLM, resized to a
$336$-pixel budget with preserved aspect ratio ($24\times24$ patches of $14$
pixels for square views), and outputs the 1024-dimensional patch features of
its final block (index $23$). After adding a two-dimensional sinusoidal
positional embedding scaled by $0.1$, the patch map is bilinearly
interpolated onto the token grid. A LayerNorm--MLP projector with hidden
width $2048$ maps layer-25 VLM tokens to $1024$ dimensions. Each view is
processed independently in bfloat16.

% 中文：接触对齐。离线生成的接触热图（附录~\ref{app:contact_maps}）在训练时双线性缩放到
% 384×384 的头部观测，按面积平均池化到 12×12 网格并做最大值归一化，作为软标签。监督只作用于
% 头部视角的第 25 层 token；LayerNorm--MLP 投影器隐藏宽度 256、输出宽度 128，后接线性预测头，
% 以式~\eqref{eq:contact_alignment} 中的 focal（指数 2）与 Dice 损失训练。
% 过滤在原始热图上进行：超过 0.05 的像素占比不少于 0.5% 时保留，全零静止帧以 0.05 的概率保留，
% 其余低于阈值的帧跳过。
\paragraph{Contact alignment}
The offline contact heatmap (Appendix~\ref{app:contact_maps}) is
bilinearly resized to the $384\times384$ head observation,
area-average-pooled onto the $12\times12$ token grid, and max-normalized
into the soft target $y_j$. Supervision uses layer-25 tokens of the head
view only. A LayerNorm--MLP projector with hidden width $256$ and output
width $128$ feeds a linear map head, trained with the focal (exponent $2$)
and Dice losses of Eq.~\eqref{eq:contact_alignment}. Frames are filtered on
the raw map:
a frame is kept when at least $0.5\%$ of its pixels exceed $0.05$,
all-zero idle frames are kept with probability $0.05$, and the remaining
sub-threshold frames are skipped.

\subsection{Contact-Map Generation}
\label{app:contact_maps}

% 中文：接触图由 320 个观测 taxel 生成，左右手分别计算。MANO 手部姿态由手套重建，因此人类
% 数据与同款手套机器人数据采用相同流程。
Contact maps are generated offline from the $320$ observed taxels of each
hand, processed separately. The MANO hand pose is reconstructed from the
glove, so human data and same-glove robot data share the same pipeline.

% 中文：接触检测。压力减去静息偏置 10 并截断为非负。噪声尺度 σ 按 taxel 在整段序列上估计：
% 跳过前 10 帧，取其后至多 60 帧为静息窗口并计算 MAD；若窗口被接触污染，则改用整段序列中最
% 安静 20% 样本的 MAD；σ 下限为 1。幅值与 24 帧升幅除以 σ 后均不小于 2 时 taxel 进入接触；
% 之后幅值不低于 1 即保持，连续 5 帧低于 1、或低于 1 且回落不小于 2σ 时退出。进入时的升幅在
% 接触期间固定。退出后强度在 8 帧内线性衰减。区域强度取其 taxel 最大值，手的强度 A 取各接触
% 区域的最大值。
\paragraph{Contact detection}
Each taxel's pressure $p$ is offset-corrected as $Q=\max(p-10,0)$. A
per-taxel noise scale $\sigma$ is estimated once per sequence: after
skipping the first $10$ frames, the next (up to) $60$ frames form a rest
window $R$ with robust scale
$\mathrm{MAD}(X)=1.4826\,\mathrm{median}(|X-\mathrm{median}(X)|)$. Let $L$
be the running 3-frame maximum of $Q$. If $R$ is contaminated by contact,
\ie $\mathrm{MAD}(R)>\max(3,\,3\max(\sigma_{\mathrm{fb}},1))$ or $\max(R)$
exceeds the 70th percentile of $L$, we use the fallback $\sigma_{\mathrm{fb}}$,
the MAD of the quietest $20\%$ of $L$; $\sigma$ is clamped to at least $1$.
With $z_{\mathrm{level}}$ the 3-frame maximum of $Q$ and $z_{\mathrm{rise}}$
the rise of the 24-frame peak over its preceding minimum, both divided by
$\sigma$, a taxel enters contact when $z_{\mathrm{level}}\geq2$ and
$z_{\mathrm{rise}}\geq2$. Contact is held while $z_{\mathrm{level}}\geq1$
and released after five consecutive frames below this floor, or immediately
when $z_{\mathrm{level}}<1$ and the pressure has fallen by at least
$2\sigma$. The entry rise $z_{\mathrm{enter}}$ is frozen for the contact
episode, and the taxel intensity is
\begin{equation}
  A_{\mathrm{taxel}}
  =
  \operatorname{clip}\!\left(\tfrac{z_{\mathrm{level}}}{6},0,1\right)
  \operatorname{clip}\!\left(0.35+0.65\,\tfrac{z_{\mathrm{enter}}}{6},0,1\right),
\end{equation}
which combines the current pressure level with the sharpness of contact
onset. After release, the intensity decays linearly to zero over eight
frames. A region's intensity is the maximum over its taxels, and the hand
intensity $A\in[0,1]$ is the maximum over the regions in contact.

% 中文：表面扩散。每个接触区域按 z_level·z_rise 选取至多 3 个接触 taxel，映射到 MANO 掌侧
% 顶点作为源（无命中时回退到该区域的顶点集合），按测地距离衰减得到 778 个顶点的亮度；无接触
% 时整只手为 0。
\paragraph{Surface diffusion}
In each region in contact, up to three active taxels with the highest
$z_{\mathrm{level}}\,z_{\mathrm{rise}}$ are mapped onto palmar-side
vertices of the 778-vertex MANO mesh, which serve as sources; if the mapping
yields no vertex, the region's vertex set is used instead. With $d(v)$ the
geodesic distance from vertex $v$ to the nearest source and
$d_{\mathrm{ref}}$ the 98th percentile of $d(v)$, clamped to at least
$0.03$~m, the vertex intensity is
\begin{equation}
  F(v)
  =
  \max\!\left(
    (0.35+0.65A)\,
    \operatorname{clip}\!\left(1-\tfrac{d(v)}{d_{\mathrm{ref}}},0,1\right),\,
    0.05
  \right),
\end{equation}
and $F\equiv0$ when the hand has no active contact.

% 中文：渲染。MANO 顶点经逐帧外参与标定内参投影到头部相机，深度小于 0.03 m 的顶点丢弃；
% 三个顶点均落在图像边界外扩 8 像素以内的三角形才绘制，按平均深度由远到近覆盖，像素用重心
% 坐标插值顶点亮度。左右手取逐像素最大值并截断到 [0,1]，得到 480×640 热图。
\paragraph{Rendering}
The posed MANO vertices are projected into the head camera with the
per-frame extrinsics and calibrated intrinsics; vertices with depth below
$0.03$~m are discarded. A triangle is rasterized only when all three
vertices project within $8$ pixels of the image border, and triangles are
painted from far to near by mean depth. Each covered pixel takes the
barycentric interpolation of its vertex intensities, with negative weights
clipped and renormalized. The two hands are merged by a pixelwise maximum
and clipped to $[0,1]$, giving a $480\times640$ map.

% 中文：目标性质。(i) 空间范围：热图覆盖几乎整只可见手，峰值位于接触源附近；结合 12×12 网格，
% 目标刻画手部层级的接触区域而非精确接触点。(ii) 强度范围：接触手的峰值位于 [0.41,1]，无接触
% 手为 0；逐帧最大值归一化只保留帧内相对分布。(iii) 过滤：过滤阈值等于亮度下限 0.05，只有接触
% 源附近区域足够大时才保留该帧。(iv) 离线性：σ 使用整段序列，属于非因果估计，但接触图只作训练
% 目标。(v) 遮挡：只处理同一只手内部的自遮挡，目标表示手在何处接触，包括被物体遮挡的区域。
\paragraph{Properties of the contact targets}
\emph{Spatial extent.} Because $d_{\mathrm{ref}}$ is the 98th percentile of
the geodesic distances and every vertex of a contacting hand receives at
least $0.05$, the map covers nearly the entire visible hand, with intensity
peaking near the sources. Together with the $12\times12$ token grid ($32$
pixels per token), the targets encode hand-level contact regions rather
than precise contact points.
\emph{Intensity range.} A taxel in contact has $z_{\mathrm{level}}\geq1$ and
$z_{\mathrm{enter}}\geq2$, so
$A\geq\tfrac{1}{6}\bigl(0.35+0.65\cdot\tfrac{2}{6}\bigr)\approx0.09$; the
peak intensity of a contacting hand therefore lies in $[0.41,1]$, and a hand
without contact contributes zero. Per-frame max-normalization retains only
the relative distribution within each frame.
\emph{Filtering.} The filtering threshold equals the intensity floor
($0.05$), so a frame is retained only when the region near the sources,
where $(0.35+0.65A)(1-d/d_{\mathrm{ref}})>0.05$, covers enough of the image.
\emph{Offline estimation.} The fallback noise scale uses the entire sequence
and is thus non-causal, which is acceptable because contact maps are
training targets only.
\emph{Occlusion.} Depth ordering resolves self-occlusion within each hand
only, so the targets mark where the hand is in contact, including regions
hidden behind the manipulated object.

\section{Training and Deployment}
\label{app:pretraining_recipe}

\subsection{Pretraining}

% 中文：完整 Spark-0 的预训练配方以 pretrain_spark_0 发布。模型基于 StarVLA 搭建，代码遵循
% XPolicyLab 标准。预训练共 400k 步，所有预训练变体使用该 checkpoint。七个数据来源及其采样概率见
% 表~\ref{tab:pretrain_mixture}，权重之和为 1；触觉存在于 HumanTouch（0.35）和同款手套机器人
% 数据（0.09）中。模型与优化超参数见表~\ref{tab:model_hparams}。
\paragraph{Recipe}
The full Spark-0 pretraining recipe is released as
\texttt{pretrain\_spark\_0}. The model is built on StarVLA
\cite{community2026starvla}, and its training, evaluation, and deployment
code follows the XPolicyLab standard \cite{community2026xpolicylab}.
Pretraining runs for 400k optimizer steps, and all pretrained variants use
the resulting checkpoint. Table~\ref{tab:pretrain_mixture} lists the seven
sources and their sampling probabilities, which sum to one; touch is present
in HumanTouch ($0.35$) and the same-glove robot data ($0.09$).
Table~\ref{tab:model_hparams} lists the model and optimization
hyperparameters.

% 中文：输入。相机打包顺序为头部、左腕、右腕；头部图像缩放至 384×384，腕部 256×256，灵活的
% 视角掩码允许腕部相机缺失。原始数据每侧存储 7 维末端位姿（xyz 与四元数）和 20 个关节角，
% 输入模型时按式~\eqref{eq:action_space} 重编码为 58 维状态与动作。状态与动作通道采用 q99
% 归一化。
\paragraph{Inputs}
Cameras are packed in the order head, left wrist, right wrist. Head images
are resized to $384\times384$ and wrist images to $256\times256$, and
flexible view masks allow missing wrist cameras. Raw data store each side
as a 7-d end-effector pose (xyz and quaternion) and $20$ joint angles, which
are re-encoded into the 58-dimensional states and actions of
Eq.~\eqref{eq:action_space}. State and action channels use $q_{99}$
normalization.

\begin{table*}[t]
  \centering
  \footnotesize
  \setlength{\tabcolsep}{5pt}
  \renewcommand{\arraystretch}{1.12}
  % 中文图注：Spark-0 预训练数据混合。数据暴露由采样权重而非时长决定。
  \caption{\textbf{Spark-0 pretraining mixture.} Exposure is set by sampling
  weight, not hours.}
  \label{tab:pretrain_mixture}
  \begin{tabular}{lrrll}
    \toprule
    \textbf{Source} & \textbf{Hours} & \textbf{Weight} & \textbf{Views}
    & \textbf{Primary role} \\
    \midrule
    HumanTouch & $1{,}000$ & 0.35 & head + wrists
      & human contact \\
    EgoDex & 662 & 0.39 & head
      & scene and motion diversity \\
    Same-glove robot play & 20 & 0.09 & robot views
      & embodiment and contact \\
    Robot play, tactile removed & 24 & 0.05 & matched views
      & scene-matched visual control \\
    Simulation & 25 & 0.05 & simulated
      & visual domain randomization \\
    Fourier ActionNet & 153 & 0.04 & head
      & robot task diversity \\
    AgiBot World & 60 & 0.03 & head + wrists
      & multi-view robot diversity \\
    \bottomrule
  \end{tabular}
\end{table*}

\begin{table*}[t]
  \centering
  % 中文图注：模型（左）与优化（右）超参数；在 12 台 8 卡节点上训练。
  \caption{\textbf{Model (left) and optimization (right) hyperparameters.}
  Training uses $12$ nodes with $8$ GPUs each.}
  \label{tab:model_hparams}
  \begingroup
  \footnotesize
  \setlength{\tabcolsep}{4pt}
  \renewcommand{\arraystretch}{1.12}
  \fitwidth{%
  \begin{tabular}{@{}ll@{\hspace{2.5em}}ll@{}}
    \toprule
    \textbf{Component} & \textbf{Setting}
    & \textbf{Parameter} & \textbf{Setting} \\
    \midrule
    VLM & Qwen3-VL-4B-Instruct; FlashAttention-2
      & Optimizer & AdamW; $\beta=(0.9,0.95)$; $\epsilon=10^{-8}$ \\
    Action queries & 64; query + visual-language KV
      & Weight decay & $10^{-8}$ \\
    Action DiT & 36 layers; width 1024; 16 heads
      & Steps / warmup & 400k / 20k \\
    DiT regularization & dropout 0.1; final dropout on
      & Devices / global batch & $12\times8$ GPUs; $576$ ($6$ per GPU), no accum. \\
    State / action & 58 dims; 3 state frames; 50-step chunk
      & Base / VLM LR & $2.5\times10^{-5}$ / $1\times10^{-5}$ \\
    Flow samples & 6 noise draws per example
      & Action, projector, tactile, alignment LR & $1\times10^{-4}$ \\
    Flow time & $\tau=(0.999-u)/0.999$, $u\!\sim\!\mathrm{Beta}(1.5,1)$
      & LR schedule & cosine; minimum $5\times10^{-7}$ \\
    Flow-time buckets & 1000
      & Gradient clipping & 1.0 \\
    Action loss & uniform velocity MSE
      & Checkpoint / eval.\ interval & 5k / 10k steps \\
    Inference & 4 Euler steps
      & Seed & 0 \\
    \bottomrule
  \end{tabular}}
  \endgroup
\end{table*}

\subsection{Fine-tuning and deployment}
\label{app:downstream}

% 中文：微调。各预训练变体从对应预训练 checkpoint 热启动并复用其归一化统计量，Scratch 则从
% Qwen3-VL 初始化。所有变体保持与预训练相同的结构、优化器、学习率分组和单卡 batch size，
% 训练 80k 步，其中前 8k 步预热。几何对齐与接触对齐保持开启，DA3 教师输入改为 392 像素。
% 配方以 sft_spark_0 发布。
\paragraph{Fine-tuning}
Each pretrained Spark-0 variant is warm-started from its pretraining
checkpoint and reuses its normalization statistics, whereas Scratch starts
from the Qwen3-VL initialization. Fine-tuning keeps the architecture,
optimizer, learning-rate groups, and per-GPU batch size of pretraining and
runs for 80k steps with an 8k-step warmup. Geometry and contact alignment
remain enabled, with the DA3 teacher input enlarged to $392$ pixels. The
recipe is released as \texttt{sft\_spark\_0}.

% 中文：部署。真机部署使用无需训练的实时分块（RTC）：当前动作块剩余 26 步时异步请求下一块
% （H=50，即每 24 步重新规划），并按测得的推理延迟（标称 8 步）以 ΠGDM 引导新块与尚未执行的
% 动作衔接。缺少触觉流时填入可学习占位符，即纯视觉部署。
\paragraph{Deployment}
Deployment uses training-free real-time chunking (RTC)~\cite{black2025rtc}.
The next chunk is requested asynchronously when $26$ actions of the current
chunk remain, so the policy replans every $24$ steps for $H=50$, and
$\Pi$GDM guidance with the measured inference delay (nominally $8$ steps)
keeps the new chunk consistent with the actions still being executed. When
a tactile stream is unavailable, the learned placeholders are inserted,
giving vision-only deployment.

\section{SparkArena Details}
\label{app:sparkarena}

\subsection{Gen 1 Prototype embodiment}
\label{app:sky1}

\begin{figure}[!htb]
  \centering
  \includegraphics[width=0.55\linewidth,trim={0 7.5 0 0},clip]{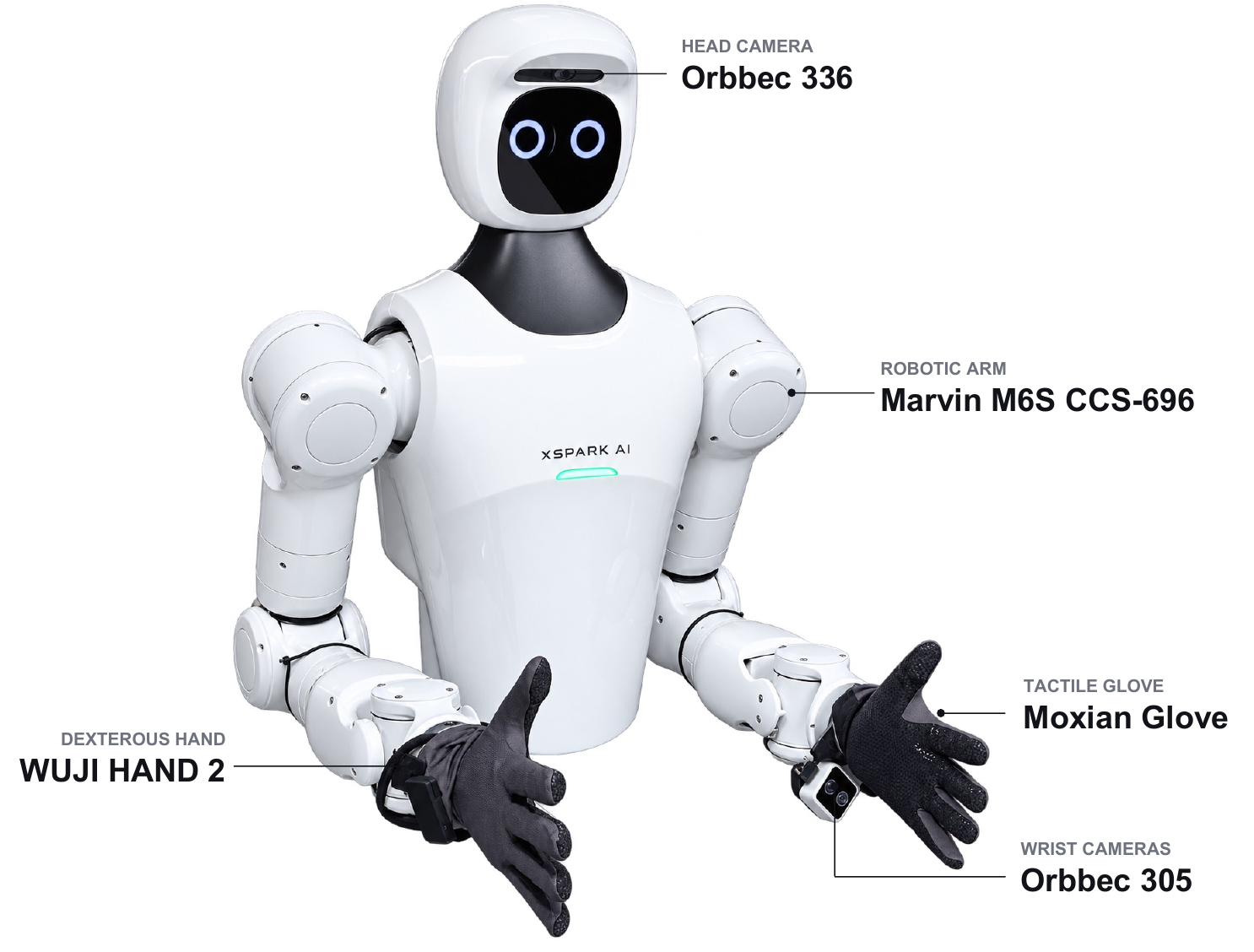}
  % 中文图注：Gen 1 Prototype 本体。真机平台在灵巧手上安装 Moxian 触觉手套。
  \caption{\textbf{Gen 1 Prototype embodiment.} The real platform mounts Moxian tactile
  gloves on the dexterous hands.}
  \label{fig:sky1-hardware}
\end{figure}

% 中文：SparkArena 的两条轨道与真机微调数据均使用 Gen 1 Prototype 本体（图~\ref{fig:sky1-hardware}）：
% 两条 Marvin M6S CCS-696 机械臂、两只 WUJI Hand 2 灵巧手、一个头部相机和两个腕内嵌相机，
% 与 HumanTouch 的视角布局一致。真机平台在灵巧手上安装 Moxian 触觉手套，采用与 HumanTouch
% 相同的 320 taxel 布局与区域划分。SparkArena-Sim 复现相同的运动学与相机配置，但不含触觉。
Both SparkArena tracks and the real-robot fine-tuning data use the Gen 1 Prototype
embodiment (Figure~\ref{fig:sky1-hardware}): two Marvin M6S CCS-696 arms,
two WUJI Hand 2 dexterous hands, a head camera, and two wrist-embedded
cameras matching the view layout of HumanTouch. The real platform mounts
Moxian tactile gloves on the hands with the same 320-taxel layout and
regional partition as HumanTouch. SparkArena-Sim reproduces the kinematics
and cameras without tactile input.

\subsection{Success and scoring criteria}
\label{app:sparkarena-scoring}

% 中文：表~\ref{tab:sparkarena-scoring} 给出每项任务的成功条件与部分得分。试验得分取其达到的
% 最高阶段分值，未达到任何阶段记 0 分；得分为 100 即为成功，无中间阶段的任务按二值计分。
Table~\ref{tab:sparkarena-scoring} defines the success condition and partial
credit of each task. A trial receives the value of the highest stage it
reaches, or $0$ if it reaches none, and succeeds exactly when it scores
$100$; tasks without intermediate stages are scored as binary success.

\begin{table*}[t]
  \centering
  % 中文表注：SparkArena 评分标准。括号内为该阶段分值；试验得分取达到的最高阶段，
  % 达到 100 即为成功。
  \caption{\textbf{SparkArena scoring criteria.} Values in parentheses are
  stage scores. A trial scores its highest reached stage and succeeds at
  $100$.}
  \label{tab:sparkarena-scoring}
  \begingroup
  \footnotesize
  \setlength{\tabcolsep}{5pt}
  \renewcommand{\arraystretch}{1.15}
  \begin{tabular}{@{}
    >{\raggedright\arraybackslash}p{0.18\textwidth}
    >{\raggedright\arraybackslash}p{0.39\textwidth}
    >{\raggedright\arraybackslash}p{0.34\textwidth}
  @{}}
    \toprule
    \textbf{Task} & \textbf{Success condition (100)}
      & \textbf{Partial credit} \\
    \midrule
    \multicolumn{3}{@{}l}{\textit{SparkArena-Sim}} \\
    Click Mouse
      & Mouse moved onto the pad and left button pressed
      & Binary \\
    Collect Objects
      & Three objects placed in the central basket
      & One object (15); two objects (40) \\
    Dual Bottles Pick
      & Both bottles lifted
      & One bottle lifted (30) \\
    Hammer Beat
      & Hammer lifted ${\geq}2$~cm above its initial height, then struck
        head-down on the block
      & Binary \\
    Put Food in Microwave
      & Food inside the microwave with the door closed
      & Binary \\
    Retrieve Gap
      & Mario figure standing upright on the gray mat
      & Binary \\
    Stack Bowls
      & Three bowls stacked
      & Two bowls stacked (15) \\
    \midrule
    \multicolumn{3}{@{}l}{\textit{SparkArena-Real}} \\
    Insert Test Tubes
      & Three tubes inserted into the rack
      & Tube picked up (20); one inserted (50); two inserted (80) \\
    Pack Shoes into Box
      & Both shoes placed in the box
      & One shoe placed (50) \\
    Place the Phone
      & Phone placed on the stand
      & Phone picked up (30) \\
    Stack Bowls
      & Three bowls stacked
      & Bowl moved to the center, if not initially there (30); two bowls
        stacked (70) \\
    Stand Up Bottle
      & Bottle standing upright
      & Binary \\
    \bottomrule
  \end{tabular}
  \endgroup
\end{table*}

\subsection{Real-robot task assets}
\label{app:sparkarena-real}

\begin{figure*}[t]
  \centering
  \includegraphics[width=\textwidth,trim={0 9 0 0},clip]{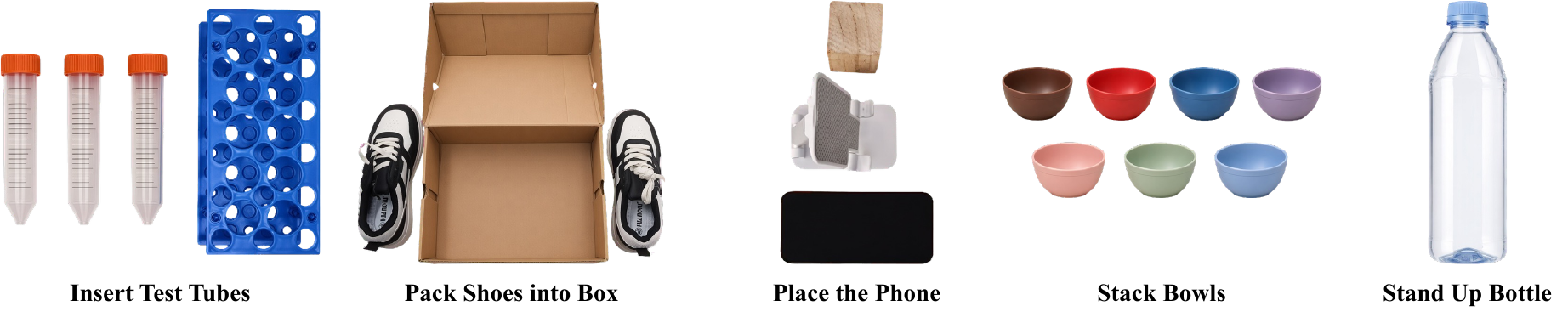}
  % 中文图注：真机任务资产。每列对应一项 SparkArena-Real 任务，展示该任务使用的
  % 物体、容器与操作目标。
  \caption{\textbf{Real-robot task assets.} Each column corresponds to one
  SparkArena-Real task and shows the objects, containers, and manipulation
  targets used in that task.}
  \label{fig:sparkarena-real-assets}
\end{figure*}

% 中文：为支持可复现评测，我们开源完整的 SparkArena-Real 设置（硬件与传感器配置、任务定义、
% 成对初始条件及评分协议）以及 SparkArena-Sim 的仿真环境与评测代码。图~\ref{fig:sparkarena-real-assets}
% 展示各真机任务所用物体，表~\ref{tab:sparkarena-real-asset-links} 给出采购链接。
To support reproducible evaluation, we release the complete SparkArena-Real
setup, including the hardware and sensor configuration, task definitions,
paired initial conditions, and scoring protocols, together with the
SparkArena-Sim environment and evaluation code.
Figure~\ref{fig:sparkarena-real-assets} shows the objects used in each
real-robot task, and Table~\ref{tab:sparkarena-real-asset-links} lists
their purchase links.

\begin{table*}[t]
  \centering
  % 中文表注：SparkArena-Real 真机任务资产采购链接。链接于 2026 年 10 月访问；
  % 同规格替代品也可用于复现实验。
  \caption{\textbf{Purchase links for SparkArena-Real task assets.}
  Accessed October 2026; equivalent items of matching size and material may
  be substituted.}
  \label{tab:sparkarena-real-asset-links}
  \begingroup
  \footnotesize
  \setlength{\tabcolsep}{8pt}
  \renewcommand{\arraystretch}{1.12}
  \begin{tabular}{@{}lll@{}}
    \toprule
    \textbf{Task} & \textbf{Asset} & \textbf{Source} \\
    \midrule
    Insert Test Tubes
      & Dual-purpose centrifuge-tube rack (ZH9112)
      & \href{https://3.cn/37-5Y2G8?jkl=@J3rGq9SgX3@}{JD} \\
      & 50-mL conical centrifuge tubes, pack of 25 (MF3390)
      & \href{https://3.cn/37-5Z7hV?jkl=@IAalABIY84@}{JD} \\
    \addlinespace
    Pack Shoes into Box
      & Black-and-white platform sneakers
      & \href{https://e.tb.cn/h.j0D54LkmpnQEOBb?tk=9TG8TsGg3gj}{Taobao} \\
    \addlinespace
    Place the Phone
      & Foldable adjustable phone stand (MF168)
      & \href{https://e.tb.cn/h.jZEX7cvgFDIfJ5j?tk=hLeDTsG3c6T}{Taobao} \\
      & Colored wooden cubes (HU006)
      & \href{https://e.tb.cn/h.j0D4eKxt2C6Tf6d?tk=x763TsG4fFs}{Taobao} \\
      & Smartphone display model (CZ321)
      & \href{https://e.tb.cn/h.jYVzEQRFCHpvKFY?tk=if1sTsGhQQE}{Taobao} \\
    \addlinespace
    Stack Bowls
      & Bowls
      & \href{https://item.jd.com/10078360050112.html}{JD} \\
    \addlinespace
    Stand Up Bottle
      & Evian natural mineral-water bottle, 500 mL (MF287)
      & \href{https://e.tb.cn/h.jZEYcMepq1ooTOH?tk=LQbpTsGX914}{Taobao} \\
    \bottomrule
  \end{tabular}
  \endgroup
\end{table*}

\subsection{Contact sequences}

\begin{figure*}[t]
  \centering
  \includegraphics[width=\textwidth,trim={0 5.5 0 0},clip]{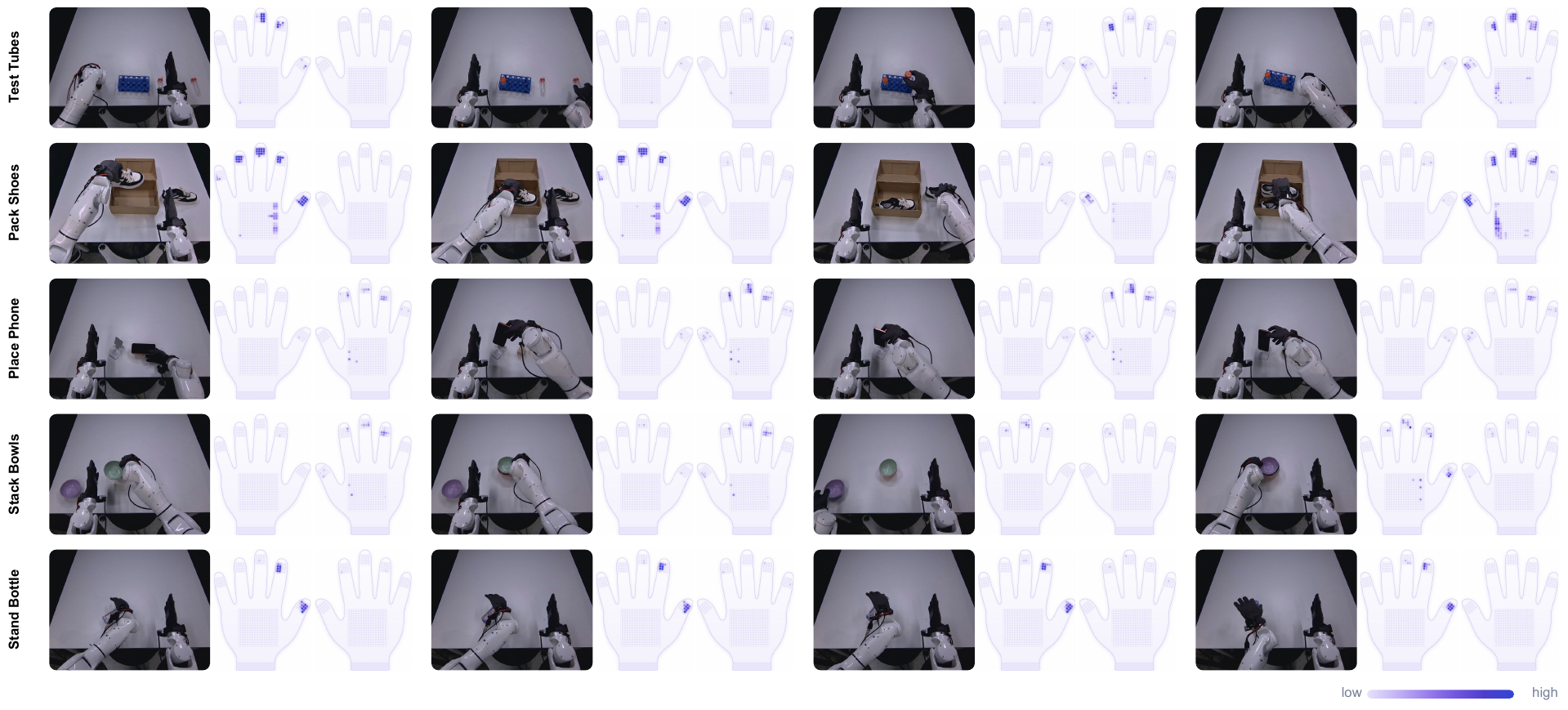}
  % 中文图注：SparkArena-Real 接触序列。每行一项任务，从左到右为触觉响应随时间的演化。
  \caption{\textbf{SparkArena-Real contact sequences.} Each row is one task;
  columns show the tactile response over time.}
  \label{fig:sparkarena-real-details}
\end{figure*}

% 中文：图~\ref{fig:sparkarena-real-details} 将图~\ref{fig:sparkarena-real} 中的每项
% SparkArena-Real 任务展开为执行过程中观测到的接触序列。
Figure~\ref{fig:sparkarena-real-details} expands each SparkArena-Real task in
Figure~\ref{fig:sparkarena-real} into the contact sequence observed during
execution.

\section{Baseline Details}
\label{app:baselines}

% 中文：基线覆盖不同的架构与数据扩展策略。通用机器人策略：π0.5、GR00T-N1.7 和 StarVLA 将
% 预训练 VLM 与 flow-matching 动作专家结合；RDT-1B 是在百万级多机器人轨迹上以统一动作空间
% 预训练的 1.2B 扩散 Transformer；X-VLA 以本体相关的 soft prompt 适配跨机器人骨干。人类先验
% 迁移：EgoVLA 从第一视角视频预测人手运动，再重定向并在机器人数据上微调；H-RDT 在带位姿标注
% 的 EgoDex 轨迹上预训练 2B flow-matching 扩散 Transformer；VITRA 将百万段人类视频转为语言
% 条件的 3D 手部动作；UniT 学习视觉锚定的离散潜动作并用 flow head 解码。其他：FastWAM 联合
% 训练视频与动作 DiT，推理时去掉视频生成；另含 ACT、多流策略 RLDX-1 与跨传感器触觉基础策略 FTP-1。
The baselines cover distinct architectural and data-scaling strategies.
\emph{Generalist robot policies.} $\pi_{0.5}$, GR00T-N1.7, and StarVLA couple
a pretrained VLM to a flow-matching action expert
\cite{intelligence2025pi_,bjorck2025gr00t,community2026starvla}. RDT-1B is a
1.2B diffusion transformer pretrained on more than one million multi-robot
trajectories with a unified action space \cite{liu2024rdt}. X-VLA adapts a
cross-robot flow-matching backbone through embodiment-specific soft prompts
\cite{zheng2025x}.
\emph{Human-prior transfer.} EgoVLA predicts human wrist and hand motion from
egocentric video, then retargets and fine-tunes on robot demonstrations
\cite{yang2025egovla}. H-RDT pretrains a 2B flow-matching diffusion
transformer on pose-annotated EgoDex trajectories \cite{bi2025h}. VITRA
converts one million human-video clips into language-conditioned 3D hand
actions \cite{li2026scalable}. UniT learns visually anchored discrete latent
actions and decodes them with a flow head \cite{chen2026unit}.
\emph{Other policies.} FastWAM co-trains video and action DiTs as a
world-action model and drops video generation at inference
\cite{yuan2026fast}. We also include ACT \cite{zhao2023learning}, the
multi-stream RLDX-1 policy \cite{kim2026rldx}, and the cross-sensor tactile
foundation policy FTP-1 \cite{yuan2026ftp}.

\section{Additional Results}
\label{app:additional_results}

\subsection{Per-task Ego Humanoid results}
\label{app:egovla_detailed}

% 中文：表~\ref{tab:egovla-short-horizon} 与表~\ref{tab:egovla-long-horizon} 给出正文汇总表
% 对应的逐任务结果，每个汇总值为对应任务的算术平均。Spark-0（1000H）四项平均成功率为 80.95、
% 75.97、68.15 和 61.82，整体平均 71.72；Visual PT 与 100H 的整体平均分别为 52.45 和 59.42。摘要与
% 引言中的 52.5% 和 71.7% 为 Visual PT 与 1000H 的四舍五入值。
Tables~\ref{tab:egovla-short-horizon} and~\ref{tab:egovla-long-horizon}
report the per-task results behind Table~\ref{tab:egovla-summary}; each
aggregate is the arithmetic mean over its tasks. For Spark-0 (1000H), the
four mean success rates are $80.95\%$, $75.97\%$, $68.15\%$, and $61.82\%$,
with an overall mean of $71.72\%$; the overall means of the Visual PT and 100H
variants are $52.45\%$ and $59.42\%$. The abstract and introduction round
the Visual PT and 1000H values to $52.5\%$ and $71.7\%$.

\begin{table*}[t]
\centering
% 中文图注：Ego Humanoid 短时序逐任务结果（seen / unseen）。
\caption{\textbf{Ego Humanoid short-horizon results} per task, seen and
unseen.}
\label{tab:egovla-short-horizon}
\begingroup
\footnotesize
\setlength{\tabcolsep}{3.5pt}
\renewcommand{\arraystretch}{1.12}
\fitwidth{%
\begin{tabular}{l *{16}{c}}
\toprule
Method
& \multicolumn{2}{c}{Stack-Can}
& \multicolumn{2}{c}{Push-Box}
& \multicolumn{2}{c}{Open-Drawer}
& \multicolumn{2}{c}{Close-Drawer}
& \multicolumn{2}{c}{Flip-Mug}
& \multicolumn{2}{c}{Pour-Balls}
& \multicolumn{2}{c}{Open-Laptop}
& \multicolumn{2}{c}{Mean} \\
\cmidrule(lr){2-3}
\cmidrule(lr){4-5}
\cmidrule(lr){6-7}
\cmidrule(lr){8-9}
\cmidrule(lr){10-11}
\cmidrule(lr){12-13}
\cmidrule(lr){14-15}
\cmidrule(lr){16-17}
& SR $\uparrow$ & PSR $\uparrow$
& SR $\uparrow$ & PSR $\uparrow$
& SR $\uparrow$ & PSR $\uparrow$
& SR $\uparrow$ & PSR $\uparrow$
& SR $\uparrow$ & PSR $\uparrow$
& SR $\uparrow$ & PSR $\uparrow$
& SR $\uparrow$ & PSR $\uparrow$
& SR $\uparrow$ & PSR $\uparrow$ \\
\midrule

\multicolumn{17}{c}{Seen Visual Configuration} \\

\midrule

EgoVLA~\cite{yang2025egovla}
& 77.78
& 77.78
& \runnerup{70.37}
& \runnerup{83.33}
& 59.26
& 81.48
& \sota{100.00}
& \sota{100.00}
& 59.26
& 59.26
& 77.78
& 92.59
& \sota{100.00}
& \sota{100.00}
& \runnerup{77.78}
& \runnerup{84.92} \\

FastWAM~\cite{yuan2026fast}
& 77.78
& 77.78
& \sota{77.78}
& \sota{88.89}
& 70.37
& 85.19
& \sota{100.00}
& \sota{100.00}
& \runnerup{88.89}
& 88.89
& 33.33
& 64.20
& 59.26
& 77.78
& 72.49
& 83.25 \\

UniT~\cite{chen2026unit}
& \sota{100.00}
& \sota{100.00}
& 62.96
& 80.25
& 48.15
& 48.15
& 55.56
& 62.96
& 59.26
& 81.48
& \runnerup{85.19}
& 92.59
& 62.96
& 64.81
& 67.72
& 75.75 \\

VITRA~\cite{li2026scalable}
& 81.48
& 93.83
& 22.22
& 56.79
& 40.74
& 40.74
& 59.26
& 75.93
& 62.96
& 82.72
& \runnerup{85.19}
& 92.59
& \runnerup{92.59}
& \runnerup{92.59}
& 63.49
& 76.46 \\

RLDX-1~\cite{kim2026rldx}
& \sota{100.00}
& \sota{100.00}
& 44.44
& 67.90
& 48.15
& 48.15
& 59.26
& 75.93
& 14.81
& 58.02
& 70.37
& 85.19
& 66.67
& 68.52
& 57.67
& 71.96 \\

GR00T-N1.7~\cite{bjorck2025gr00t}
& 74.07
& 90.12
& 14.81
& 45.68
& 48.15
& 51.85
& 70.37
& 85.19
& 40.74
& 70.37
& 81.48
& 90.74
& 59.26
& 59.26
& 55.56
& 70.46 \\

H-RDT~\cite{bi2025h}
& 70.37
& 88.89
& 37.04
& 60.49
& 33.33
& 33.33
& 37.04
& 61.11
& 40.74
& 74.07
& \runnerup{85.19}
& 92.59
& 59.26
& 59.26
& 51.85
& 67.11 \\

$\pi_{0.5}$~\cite{intelligence2025pi_}
& 92.59
& 97.53
& 22.22
& 71.60
& 18.52
& 18.52
& 40.74
& 53.70
& 18.52
& 41.98
& 55.56
& 77.78
& 77.78
& 77.78
& 46.56
& 62.70 \\

StarVLA~\cite{community2026starvla}
& 18.52
& 22.22
& 66.67
& \runnerup{83.33}
& 55.56
& 72.84
& \runnerup{88.89}
& \runnerup{95.06}
& 37.04
& 37.04
& 0.00
& 13.58
& 37.04
& 59.26
& 43.39
& 54.76 \\

RDT-1B~\cite{liu2024rdt}
& 70.37
& 85.19
& 22.22
& 70.37
& 7.41
& 9.26
& 14.81
& 22.22
& 37.04
& 67.90
& 33.33
& 66.67
& 40.74
& 40.74
& 32.28
& 51.76 \\

\midrule

Spark-0 (Scratch)
& \runnerup{96.30}
& \runnerup{98.77}
& 59.26
& 81.48
& 3.70
& 3.70
& 40.74
& 57.41
& 70.37
& \runnerup{90.12}
& \sota{92.59}
& \runnerup{96.30}
& 77.78
& 77.78
& 62.96
& 72.22 \\

Spark-0 (Visual PT)
& 81.48
& 92.59
& 33.33
& 62.96
& 48.15
& 48.15
& 59.26
& 72.22
& \sota{92.59}
& \sota{97.53}
& 55.56
& 77.78
& 85.19
& 85.19
& 65.08
& 76.63 \\

Spark-0 (100H)
& 92.59
& 96.30
& 48.15
& 79.01
& \sota{88.89}
& \runnerup{88.89}
& 48.15
& 62.96
& 70.37
& \runnerup{90.12}
& 40.74
& 70.37
& 85.19
& 85.19
& 67.72
& 81.83 \\

Spark-0 (1000H)
& \sota{100.00}
& \sota{100.00}
& 59.26
& 79.63
& \runnerup{77.78}
& \sota{90.12}
& \sota{100.00}
& \sota{100.00}
& 66.67
& 66.67
& \sota{92.59}
& \sota{97.53}
& 70.37
& 83.33
& \sota{80.95}
& \sota{88.18} \\

\midrule

\multicolumn{17}{c}{Unseen Visual Configuration} \\

\midrule

EgoVLA~\cite{yang2025egovla}
& 62.12
& 62.12
& \sota{75.76}
& \sota{84.85}
& 50.00
& \runnerup{75.25}
& \sota{98.48}
& \runnerup{98.48}
& 30.77
& 30.77
& \sota{83.33}
& \sota{94.44}
& \sota{83.33}
& \sota{87.88}
& \runnerup{69.11}
& 76.26 \\

UniT~\cite{chen2026unit}
& \sota{100.00}
& \sota{100.00}
& 53.03
& 77.78
& 39.39
& 39.39
& 66.67
& 71.97
& 53.03
& 74.75
& 71.21
& 83.33
& 48.48
& 49.24
& 61.69
& 70.92 \\

VITRA~\cite{li2026scalable}
& 57.58
& 81.82
& 15.15
& 55.56
& 48.48
& 48.48
& 68.18
& 82.58
& \runnerup{65.15}
& \runnerup{84.85}
& 80.30
& 89.39
& \runnerup{80.30}
& 81.06
& 59.31
& 74.82 \\

GR00T-N1.7~\cite{bjorck2025gr00t}
& \runnerup{93.94}
& \runnerup{97.98}
& 10.61
& 40.40
& 57.58
& 60.61
& 62.12
& 74.24
& 24.24
& 57.07
& 75.76
& 87.88
& 65.15
& 65.15
& 55.63
& 69.05 \\

RLDX-1~\cite{kim2026rldx}
& \runnerup{93.94}
& 97.47
& 27.27
& 56.57
& 45.45
& 46.97
& 78.79
& 85.61
& 3.03
& 43.43
& 59.09
& 79.55
& 68.18
& 68.94
& 53.68
& 68.36 \\

StarVLA~\cite{community2026starvla}
& 21.21
& 21.97
& 48.48
& 74.24
& 54.55
& 74.75
& 83.33
& 93.94
& 24.24
& 24.24
& 0.00
& 11.11
& 51.52
& 62.12
& 40.48
& 51.77 \\

H-RDT~\cite{bi2025h}
& 56.06
& 76.26
& 33.33
& 58.59
& 28.79
& 30.30
& 27.27
& 46.21
& 16.67
& 55.05
& 51.52
& 75.00
& 30.30
& 31.06
& 34.85
& 53.21 \\

RDT-1B~\cite{liu2024rdt}
& 74.24
& 84.85
& 27.27
& 70.71
& 12.12
& 13.64
& 25.76
& 34.85
& 33.33
& 64.65
& 34.85
& 57.58
& 31.82
& 32.58
& 34.20
& 51.26 \\

$\pi_{0.5}$~\cite{intelligence2025pi_}
& 81.82
& 89.90
& 24.24
& 65.66
& 15.15
& 15.15
& 36.36
& 46.97
& 9.09
& 31.82
& 30.30
& 65.15
& 42.42
& 42.42
& 34.20
& 51.01 \\

FastWAM~\cite{yuan2026fast}
& 22.73
& 22.73
& 12.12
& 42.42
& 27.27
& 59.09
& \runnerup{86.36}
& 87.37
& 37.88
& 37.88
& 9.09
& 28.28
& 31.82
& 40.15
& 32.47
& 45.42 \\

\midrule

Spark-0 (Scratch)
& 84.85
& 91.92
& 48.48
& 72.73
& 12.12
& 12.12
& 39.39
& 47.73
& 42.42
& 72.73
& \runnerup{81.82}
& 90.91
& 75.76
& 75.76
& 54.98
& 66.27 \\

Spark-0 (Visual PT)
& 75.76
& 88.89
& 24.24
& 64.14
& 60.61
& 60.61
& 71.21
& 80.30
& \sota{74.24}
& \sota{91.41}
& 42.42
& 71.21
& 74.24
& 74.24
& 60.39
& 75.83 \\

Spark-0 (100H)
& 81.82
& 92.93
& 37.88
& 75.76
& \sota{74.24}
& 74.24
& 57.58
& 68.18
& 53.03
& 83.84
& 36.36
& 68.18
& 72.73
& 72.73
& 59.09
& \runnerup{76.55} \\

Spark-0 (1000H)
& 80.30
& 80.30
& \runnerup{66.67}
& \runnerup{83.33}
& \runnerup{72.73}
& \sota{87.37}
& \sota{98.48}
& \sota{99.49}
& 57.58
& 57.58
& 78.79
& \runnerup{92.93}
& 77.27
& \runnerup{85.61}
& \sota{75.97}
& \sota{83.80} \\

\bottomrule
\end{tabular}}
\endgroup
\end{table*}
\begin{table*}[t]
\centering
% 中文图注：Ego Humanoid 长时序逐任务结果（seen / unseen）。
\caption{\textbf{Ego Humanoid long-horizon results} per task, seen and
unseen.}
\label{tab:egovla-long-horizon}
\begingroup
\footnotesize
\setlength{\tabcolsep}{3.5pt}
\renewcommand{\arraystretch}{1.12}
\fitwidth{%
\begin{tabular}{l *{12}{c}}
\toprule
Method
& \multicolumn{2}{c}{Insert-And-Unload-Cans}
& \multicolumn{2}{c}{Stack-Can-Into-Drawer}
& \multicolumn{2}{c}{Sort-Cans}
& \multicolumn{2}{c}{Unload-Cans}
& \multicolumn{2}{c}{Insert-Cans}
& \multicolumn{2}{c}{Mean} \\
\cmidrule(lr){2-3}
\cmidrule(lr){4-5}
\cmidrule(lr){6-7}
\cmidrule(lr){8-9}
\cmidrule(lr){10-11}
\cmidrule(lr){12-13}
& SR $\uparrow$ & PSR $\uparrow$
& SR $\uparrow$ & PSR $\uparrow$
& SR $\uparrow$ & PSR $\uparrow$
& SR $\uparrow$ & PSR $\uparrow$
& SR $\uparrow$ & PSR $\uparrow$
& SR $\uparrow$ & PSR $\uparrow$ \\
\midrule

\multicolumn{13}{c}{Seen Visual Configuration} \\

\midrule

FastWAM~\cite{yuan2026fast}
& 22.22
& 73.33
& 0.00
& 39.81
& 62.96
& 90.74
& \runnerup{88.89}
& 92.59
& \sota{74.07}
& \sota{93.52}
& 49.63
& 78.00 \\

EgoVLA~\cite{yang2025egovla}
& 44.44
& \runnerup{77.04}
& 40.74
& 75.93
& 55.56
& 88.89
& 66.67
& 83.33
& 22.22
& \runnerup{78.70}
& 45.93
& \runnerup{80.78} \\

VITRA~\cite{li2026scalable}
& \sota{59.26}
& \sota{86.67}
& 7.41
& 39.81
& \runnerup{81.48}
& \runnerup{93.52}
& 18.52
& 58.33
& 33.33
& \runnerup{78.70}
& 40.00
& 71.41 \\

$\pi_{0.5}$~\cite{intelligence2025pi_}
& 0.00
& 16.67
& 29.63
& 57.41
& 14.81
& 47.22
& 40.74
& 74.07
& 11.11
& 52.59
& 19.26
& 49.59 \\

StarVLA~\cite{community2026starvla}
& 18.52
& 68.89
& 0.00
& 18.52
& 33.33
& 74.07
& 0.00
& 41.67
& 33.33
& 60.19
& 17.04
& 52.67 \\

H-RDT~\cite{bi2025h}
& 22.22
& 41.67
& 18.52
& 41.67
& 29.63
& 79.63
& 0.00
& 67.59
& 3.70
& 40.74
& 14.81
& 54.26 \\

RDT-1B~\cite{liu2024rdt}
& 0.00
& 23.15
& 11.11
& 51.85
& 25.93
& 57.41
& 3.70
& 72.22
& 11.11
& 51.85
& 10.37
& 51.30 \\

RLDX-1~\cite{kim2026rldx}
& 25.93
& 56.48
& 0.00
& 29.63
& 7.41
& 41.67
& 0.00
& 46.30
& 0.00
& 28.15
& 6.67
& 40.44 \\

UniT~\cite{chen2026unit}
& 3.70
& 56.30
& 3.70
& 34.26
& 0.00
& 55.56
& 3.70
& 33.33
& 14.81
& 43.52
& 5.19
& 44.59 \\

GR00T-N1.7~\cite{bjorck2025gr00t}
& 0.00
& 45.19
& 3.70
& 30.56
& 0.00
& 49.07
& 0.00
& 26.85
& 0.00
& 34.26
& 0.74
& 37.19 \\

\midrule

Spark-0 (Scratch)
& \runnerup{51.85}
& 63.89
& 44.44
& 74.07
& 74.07
& \runnerup{93.52}
& \runnerup{88.89}
& \runnerup{97.22}
& 48.15
& 74.81
& \runnerup{61.48}
& 80.70 \\

Spark-0 (Visual PT)
& 14.81
& 50.00
& 74.07
& 87.96
& 25.93
& 39.81
& \runnerup{88.89}
& \runnerup{97.22}
& 40.74
& 56.30
& 48.89
& 66.26 \\

Spark-0 (100H)
& 37.04
& 58.33
& \runnerup{81.48}
& \sota{92.59}
& 40.74
& 58.33
& \sota{100.00}
& \sota{100.00}
& 22.22
& 51.11
& 56.30
& 72.07 \\

Spark-0 (1000H)
& 29.63
& 54.81
& \sota{88.89}
& \runnerup{91.67}
& \sota{85.19}
& \sota{96.30}
& 70.37
& 87.04
& \runnerup{66.67}
& 75.00
& \sota{68.15}
& \sota{80.96} \\

\midrule

\multicolumn{13}{c}{Unseen Visual Configuration} \\

\midrule

VITRA~\cite{li2026scalable}
& \sota{59.09}
& \sota{87.58}
& 16.67
& 46.59
& 57.58
& 87.88
& 16.67
& 63.26
& \runnerup{48.48}
& \sota{85.61}
& 39.70
& 74.18 \\

EgoVLA~\cite{yang2025egovla}
& 31.82
& \runnerup{76.97}
& 28.79
& 60.98
& 18.18
& 68.94
& 50.00
& 75.76
& 15.15
& 62.88
& 28.79
& 69.11 \\

FastWAM~\cite{yuan2026fast}
& 19.70
& 47.88
& 0.00
& 8.33
& 16.67
& 40.91
& 30.30
& 51.52
& 39.39
& 47.73
& 21.21
& 39.27 \\

StarVLA~\cite{community2026starvla}
& 19.70
& 71.52
& 3.03
& 17.05
& 24.24
& 70.08
& 4.55
& 42.80
& 33.33
& 57.58
& 16.97
& 51.80 \\

$\pi_{0.5}$~\cite{intelligence2025pi_}
& 0.00
& 14.77
& 28.79
& 51.89
& 19.70
& 45.83
& 18.18
& 67.42
& 13.64
& 51.52
& 16.06
& 46.29 \\

RDT-1B~\cite{liu2024rdt}
& 7.58
& 26.36
& 13.64
& 50.76
& 19.70
& 59.85
& 3.03
& 68.18
& 10.61
& 57.58
& 10.91
& 52.55 \\

H-RDT~\cite{bi2025h}
& 28.79
& 39.39
& 3.03
& 35.98
& 12.12
& 65.91
& 0.00
& 51.14
& 1.52
& 38.48
& 9.09
& 46.18 \\

UniT~\cite{chen2026unit}
& 10.61
& 53.03
& 4.55
& 37.12
& 0.00
& 50.00
& 4.55
& 37.50
& 12.12
& 52.65
& 6.36
& 46.06 \\

RLDX-1~\cite{kim2026rldx}
& 27.27
& 54.55
& 0.00
& 29.55
& 1.52
& 42.05
& 0.00
& 43.56
& 1.52
& 29.39
& 6.06
& 39.82 \\

GR00T-N1.7~\cite{bjorck2025gr00t}
& 1.52
& 49.09
& 7.58
& 39.77
& 0.00
& 47.35
& 0.00
& 31.82
& 0.00
& 34.85
& 1.82
& 40.58 \\

\midrule

Spark-0 (Scratch)
& 40.91
& 54.17
& 33.33
& 64.02
& \sota{86.36}
& \sota{93.94}
& \sota{63.64}
& \runnerup{86.74}
& 45.45
& 79.39
& 53.94
& 75.65 \\

Spark-0 (Visual PT)
& 12.12
& 50.00
& 53.03
& 78.41
& 21.21
& 31.82
& \runnerup{62.12}
& \sota{90.53}
& 28.79
& 53.33
& 35.45
& 60.82 \\

Spark-0 (100H)
& \runnerup{57.58}
& 75.00
& \runnerup{74.24}
& \runnerup{89.02}
& 45.45
& 70.83
& \runnerup{62.12}
& \sota{90.53}
& 33.33
& 56.67
& \runnerup{54.55}
& \runnerup{76.41} \\

Spark-0 (1000H)
& 31.82
& 53.03
& \sota{86.36}
& \sota{92.05}
& \runnerup{68.18}
& \runnerup{92.05}
& 60.61
& 81.82
& \sota{62.12}
& \runnerup{83.71}
& \sota{61.82}
& \sota{80.53} \\

\bottomrule
\end{tabular}}
\endgroup
\end{table*}

\subsection{Complete SparkArena-Real results}
\label{app:real_all}

% 中文：表~\ref{tab:real-bench-all} 给出全部基线与 Spark-0 变体在 SparkArena-Real 上的逐任务
% 结果，正文表~\ref{tab:real-bench} 与表~\ref{tab:spark0-ablation} 中的真机数值均取自此表。
Table~\ref{tab:real-bench-all} reports per-task SparkArena-Real results for
all baselines and Spark-0 variants; the real-robot entries in
Tables~\ref{tab:real-bench} and~\ref{tab:spark0-ablation} are taken from
this table.

\begin{table*}[t]
\centering
% 中文图注：SparkArena-Real 完整结果（成功率 % / 完成分数）。
\caption{\textbf{Complete SparkArena-Real results} (SR \% / completion
score).}
\label{tab:real-bench-all}
\begingroup
\footnotesize
\setlength{\tabcolsep}{3.5pt}
\renewcommand{\arraystretch}{1.12}
\fitwidth{%
\begin{tabular}{l *{6}{c}}
\toprule
\textbf{Method}
& \makecell{\textbf{Insert}\\\textbf{Test Tubes}}
& \makecell{\textbf{Pack Shoes}\\\textbf{into Box}}
& \makecell{\textbf{Place}\\\textbf{the Phone}}
& \makecell{\textbf{Stack}\\\textbf{Bowls}}
& \makecell{\textbf{Stand Up}\\\textbf{Bottle}}
& \textbf{Mean} \\
\midrule
H-RDT~\cite{bi2025h}
& 0.00 / 0.00 & 0.00 / 0.00 & 0.00 / 0.00
& 0.00 / 0.00 & 0.00 / 0.00 & 0.00 / 0.00 \\
RDT-1B~\cite{liu2024rdt}
& 0.00 / 0.00 & 0.00 / 0.00 & 0.00 / 0.00
& 0.00 / 0.00 & 0.00 / 0.00 & 0.00 / 0.00 \\
$\pi_{0.5}$~\cite{intelligence2025pi_}
& 0.00 / 8.00 & 10.00 / 50.00 & 0.00 / 15.00
& \runnerup{50.00} / \runnerup{73.00} & 10.00 / 10.00 & 14.00 / 31.20 \\
GR00T-N1.7~\cite{bjorck2025gr00t}
& 0.00 / 7.00 & 10.00 / 30.00 & 0.00 / 12.00
& 20.00 / 51.00 & 50.00 / 50.00 & 16.00 / 30.00 \\
X-VLA~\cite{zheng2025x}
& 0.00 / 17.00 & 10.00 / 10.00 & 10.00 / 10.00
& 30.00 / 54.00 & 60.00 / 60.00 & 22.00 / 30.20 \\
StarVLA~\cite{community2026starvla}
& 0.00 / 0.00 & 30.00 / 55.00 & 10.00 / 10.00
& 20.00 / 43.00 & 30.00 / 30.00 & 18.00 / 27.60 \\
FTP-1~\cite{yuan2026ftp}
& 0.00 / 24.00 & \sota{80.00} / \runnerup{80.00} & 20.00 / 32.00
& 30.00 / 53.00 & 50.00 / 50.00 & 36.00 / 47.80 \\
\midrule
Spark-0 (Scratch)
& 0.00 / \sota{35.00} & \sota{80.00} / \sota{85.00} & 20.00 / 32.00
& \sota{60.00} / \sota{77.00} & 30.00 / 30.00 & 38.00 / 51.80 \\
Spark-0 (Visual PT)
& 0.00 / \runnerup{32.00} & \sota{80.00} / \sota{85.00}
& \runnerup{40.00} / \sota{58.00}
& 20.00 / 54.00 & \sota{80.00} / \sota{80.00}
& \runnerup{44.00} / \runnerup{61.80} \\
Spark-0 (100H)
& 0.00 / 26.00 & \runnerup{50.00} / 60.00
& \runnerup{40.00} / 40.00
& 40.00 / 64.00 & \runnerup{70.00} / \runnerup{70.00} & 40.00 / 52.00 \\
\midrule
Spark-0 (1000H)
& 0.00 / 31.00 & \sota{80.00} / \sota{85.00}
& \sota{50.00} / \runnerup{56.00}
& \runnerup{50.00} / 62.00 & \sota{80.00} / \sota{80.00}
& \sota{52.00} / \sota{62.80} \\
\bottomrule
\end{tabular}}
\endgroup
\end{table*}

\end{document}